\documentclass[11pt]{article}

\usepackage{geometry}
\usepackage[font=small,labelfont=bf]{caption}

\usepackage{cite}

\usepackage{amsmath}
\usepackage{amsfonts}
\usepackage{bm}
\usepackage{graphicx}
\usepackage{xcolor}
\usepackage{float}
\usepackage{algorithm}
\usepackage{algpseudocode}
\usepackage{hyperref}
\usepackage[all]{hypcap}

\usepackage{authblk}

\usepackage{ulem} 
\usepackage{color,soul} 

\providecommand{\keywords}[1]
{
  \small	
  \textbf{\textit{Keywords --}} #1
}

\begin{document}

\pagecolor{white}

\title{Inside Out: Transforming Images of Lab-Grown Plants for \\ Machine Learning Applications in Agriculture}

\author[1,2]{Alexander E. Krosney}
\author[3]{Parsa Sotoodeh}
\author[3,\thanks{Corresponding Author. E-mail address: ch.henry@uwinnipeg.ca}]{Christopher J. Henry}
\author[3]{Michael A. Beck}
\author[2,3]{Christopher P. Bidinosti}

\affil[1]{Department of Computer Science, University of Manitoba, 66 Chancellors Circle, Winnipeg, Manitoba, Canada R3T 2N2}
\affil[2]{Department of Physics, University of Winnipeg, 515 Portage Avenue, Winnipeg, Manitoba, Canada R3B 2E9}
\affil[3]{Department of Applied Computer Science, University of Winnipeg, 515 Portage Avenue, Winnipeg, Manitoba, Canada R3B 2E9}

\date{\today}

\maketitle

\vspace{-3em}
\begin{abstract}
Machine learning tasks often require a significant amount of training data for the resultant network to perform suitably for a given problem in any domain. In agriculture, dataset sizes are further limited by phenotypical differences between two plants of the same genotype, often as a result of differing growing conditions. Synthetically-augmented datasets have shown promise in improving existing models when real data is not available. In this paper, we employ a contrastive unpaired translation (CUT) generative adversarial network (GAN) and simple image processing techniques to translate indoor plant images to appear as field images. While we train our network to translate an image containing only a single plant, we show that our method is easily extendable to produce multiple-plant field images. Furthermore, we use our synthetic multi-plant images to train several YoloV5 nano object detection models to perform the task of plant detection and measure the accuracy of the model on real field data images. Including training data generated by the CUT-GAN leads to better plant detection performance compared to a network trained solely on real data.
\end{abstract}


\section{Introduction}
\label{introduction}

Machine learning (ML) tasks are often limited by the availability and quality of training data for a given model (Goodfellow et al., 2016; LeCun et al., 2015). To enable ML-based applications in agriculture -- such as the automatic detection and classification of plants or crop health monitoring, say --  
ultimately requires large quantities of labeled image data with which to train deep neural networks (DNNs) (Liakos et al., 2018; Lobet, 2017; Lu et al., 2022; Waldchen et al., 2018). It is the present lack of such data, and the challenge of generating it, that may ultimately limit the broad application of such techniques 
across the immense variety of crop plants. Lobet (2017), for example, emphasizes that this process ``is hampered by the difficulty of finding good-quality ground-truth datasets." Similar sentiments are echoed in general reviews (Liakos et al., 2018; Lu et al., 2022; Waldchen et al., 2018), as well as in publications on specific applications such as weed detection (Bah et al., 2018; Binch and Fox, 2017; Bosilj et al., 2018) and high-throughput phenotyping (Fahlgen et al., 2015; Gehan and Kellogg, 2017; Giuffrida et al., 2018; Shakoor et al., 2017; Singh et al., 2016; Tardieu et al., 2017). The difficulty of generating or collecting plant-based 
image data for agricultural applications is further exacerbated by the many differences in growing conditions and physical dissimilarities between any two plants, even for those belonging to the same genotype.  Covering a wide variety of phenotypes with a sufficient volume of labeled training data is a task of massive scope. This challenge is further impeded by the requirement of expert knowledge that is often necessary to accurately label plant data (for example, when it comes to the distinction between oats and wild oats) (Beck et al., 2021). 
 \medbreak

Image transformation and synthesis through the use of generative adversarial networks (GANs) is gaining interest in agriculture as means to expedite the development of large-scale, balanced and ground-truthed  datasets (Lu et al., 2022).
GANs were originally used for creating synthetic MNIST digits, human faces, and other image types (Goodfellow et al., 2014) and have proven to be useful for image translation problems such as a horse to/from zebra, dog to/from cat, and summer to/from winter (Isola et al., 2016; Park et al., 2020; Zhu et al., 2017).
In the agricultural domain, GANs have been applied to areas such as plant health, weed control, and phenotyping (Lu et al., 2022).  Some specific examples that demonstrate the promise of GANs in agriculture include disease detection in leaves (Cap et al., 2020; Zeng et al., 2020), leaf counting (Giuffrida et al., 2017; Kuznichov et al., 2019; Zhu et al., 2018), modelling of seedlings (Madsen et al., 2019), and rating plant vigor (Zhu et al., 2020).

The focus of this work is not a particular task or application, but rather the development of a means to translate real indoor images of plants to appear in field settings.
This approach can in principle be used  to create bespoke, labelled data sets to support the training needs of a wide variety of ML tasks in agriculture.  For example, appropriately constructed collages of two or more species on a soil background could ultimately be used to synthesize very large numbers of images of crop plants interspersed with weeds, resulting in labelled, ground-truthed datasets suitable for developing  ML models for automated weed detection.  For this initial study, however, we limit testing to simpler demonstrations  of object detection.

The impetus for this work comes from our previous development of an embedded system for the automated generation of labeled plant images taken indoors (Beck et al., 2020).  Here, a camera mounted a computer controlled gantry system is used to take photographs of  plants against blue keying fabric  from multiple positions and angles.
Because the camera and plant positions are always known, single-plant images can be automatically cropped and labelled.  In addition to this,  we have collected outdoor images of plants and soil (Beck et al., 2021), which at present must be cropped and annotated by hand.  Belonging to both datasets are four crop species: canola, oat, soybean, and wheat. 
The presence of these plants in both datasets provides the opportunity for outdoor image synthesis through image-to-image translation via GANs (Isola et al., 2016). 
Our goal, then, is to create fully labeled training datasets 
that are visually consistent with real field data. This procedure eliminates the need for manual labeling of outdoor grown plants, which is time-consuming and prone to error, while being scalable and adaptable to new environments (e.g., different soil backgrounds, plant varieties, or weather conditions). The creation of one's own datasets could improve the accuracy of plant detection and other models in a real field setting.

This paper is structured as follows. 
Section~\ref{methods} 
provides an overview of the GAN architecture used in our image translation experiments and describes the construction process for the GAN training datasets. In
Section~\ref{results} 
we provide visual results for several single-plant translation experiments and discuss the benefits and limitations of each training dataset. 
Section~\ref{multi_plant_synthesis} 
describes our method for producing augmented outdoor multi-plant images with automated labeling. We provide plant detection results in 
Section~\ref{plant_detection} 
using a YoloV5 nano model trained on our augmented datasets.
Section~\ref{conclusion} 
concludes the paper and discusses potential extensions to the image synthesis methods.
\medbreak

\section{Methods}
\label{methods}

\begin{figure}[H]
\centering
\includegraphics[width=\linewidth]{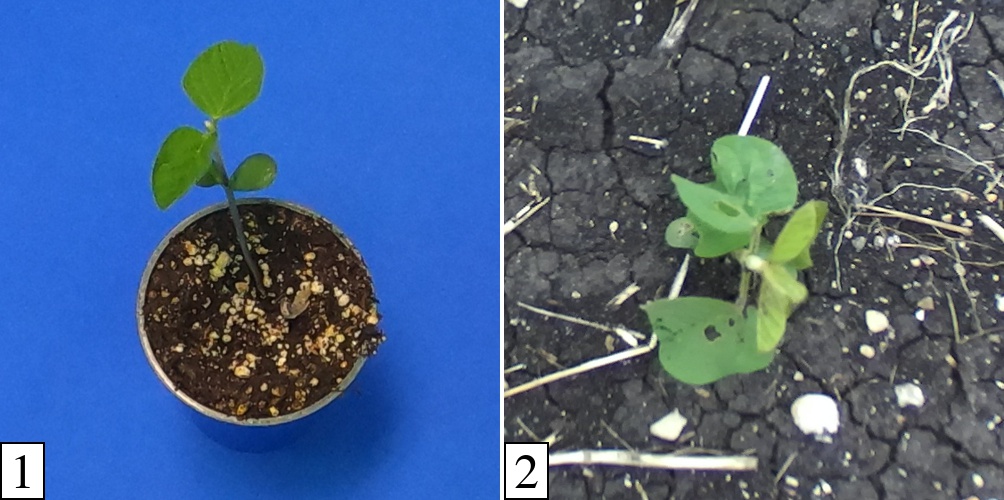}
\caption{
Image of soybean plants in indoor (1) and outdoor (2) settings. 
Indoor plants are photographed against blue keying fabric
 to enable background removal. Outdoor plants are more susceptible to leaf damage.
Differences in lighting lead to a darker appearing leaf color in the indoor images.
 }
\label{domain_examples}
\end{figure}

Many GAN architectures require the availability of paired data for training. In our case, an image pair would consist of a plant placed in front of a blue screen and an identical plant, in the same location of the image, placed in soil. Such image pairs are difficult to obtain in large volumes and instead we focus on GAN architectures that can train on unpaired data (Zhu et al., 2017), such as the examples shown in
Figure~\ref{domain_examples},
which are taken from our indoor and outdoor plant datasets.
\footnote{At the time of writing, the indoor dataset contains over 1.2 million labelled images of 14 different crops and weeds commonly found in Manitoba, Canada, while the outdoor dataset contains 540000 still images extracted from video footage of five different common crops of this region. The datasets can be made available upon request.}

Additionally, some GANs, such as CycleGAN, consist of two-sided networks that not only translate an image from one domain to another but perform the reverse translation as well (Zhu et al., 2017). For our specific problem, we are interested in single-directional translation and only consider one-sided networks to reduce training duration and model sizes. 
As a result, we follow the approach of contrastive unpaired translation (CUT) as presented in Park et al. (2020). For this, one considers two image domains $X$ and $Y$  (with samples $\{ \bm{x}_i \}_{i=1}^{N_x} \subseteq X$ and $ \{ \bm{y}_j \}_{j=1}^{N_y} \subseteq Y$) and seeks to find a function that takes a sample from the domain $X$ and outputs an image that can plausibly come from the distribution $Y$. \medbreak

\begin{figure}[H]
\centering
\includegraphics[width=\linewidth, height=0.5\linewidth]{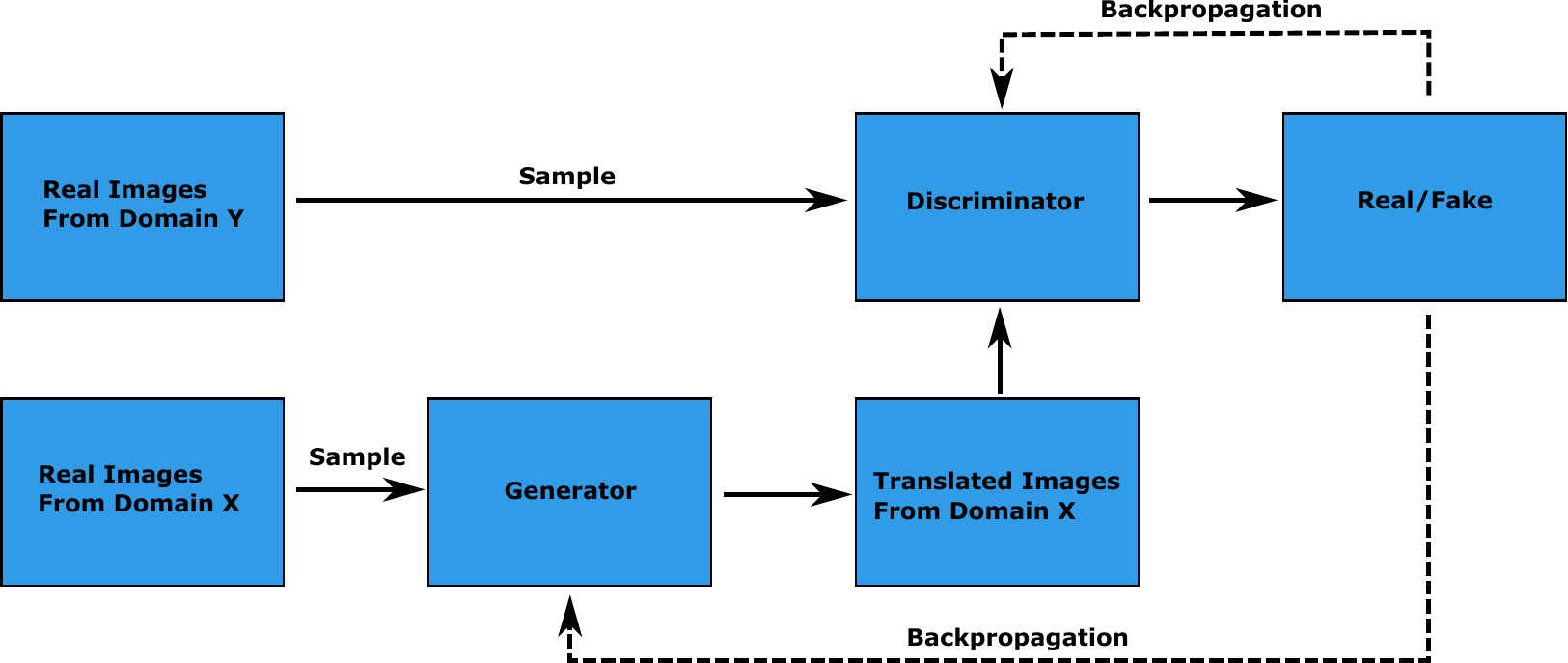}
\caption{Visual representation of an image translating GAN. The discriminator learns to differentiate between the real images from domain $Y$ and the translated images from the generator output. The discriminator and generator weights are updated through backpropagation of the gradients from the discriminator output according to the first and second terms in the adversarial loss, respectively. The image is adapted from the Google Developers Website (2022).}
\label{gan}
\end{figure}

Generative adversarial networks typically consist of two separate networks that are trained simultaneously. The generator $G$ learns the mapping $G : X \rightarrow Y$ and the discriminator $D$ is trained to differentiate between the real images of domain $Y$ and the fake images $G(\bm{x}) = \bm{\hat{y}}$ produced by the generator. Note that we refer to the real and fake images of domain $Y$ as $\bm{y}$ and $\bm{\hat{y}}$, respectively. The discriminator returns a probability in $[0.0, 1.0]$ that the input image came from the distribution $Y$. Effectively, the generator is trained to produce images that fool the discriminator by minimizing the adversarial loss (Giuffrida et al., 2017; Goodfellow et al., 2014; Isola et al., 2016; Park et al., 2020; Zhu et al., 2017; Zhu et al., 2018)
\begin{equation}
\mathcal{L}_{GAN} (G, D, X, Y) := \mathbb{E}_{\bm{y} \sim Y} \log{D(\bm{y})} + \mathbb{E}_{\bm{x} \sim X} \log{(1 - D(G(\bm{x})))}.
\end{equation}
A visual overview of a GAN structure is given in Figure~\ref{gan}.

The generator is composed of two networks that are applied sequentially to an image. The first half, the encoder $G_{enc}$, receives the input and constructs a feature stack, primarily through down-sampling operations. The second half, the decoder $G_{dec}$, takes a feature stack and constructs a new image, through up-sampling operations. Here, we have that $G(\bm{x}) = G_{dec}(G_{enc}(\bm{x})) = \bm{\hat{y}}$ (Park et al., 2020). \medbreak

\begin{figure}[H]
\centering
\includegraphics[width=\linewidth, height=0.6\linewidth]{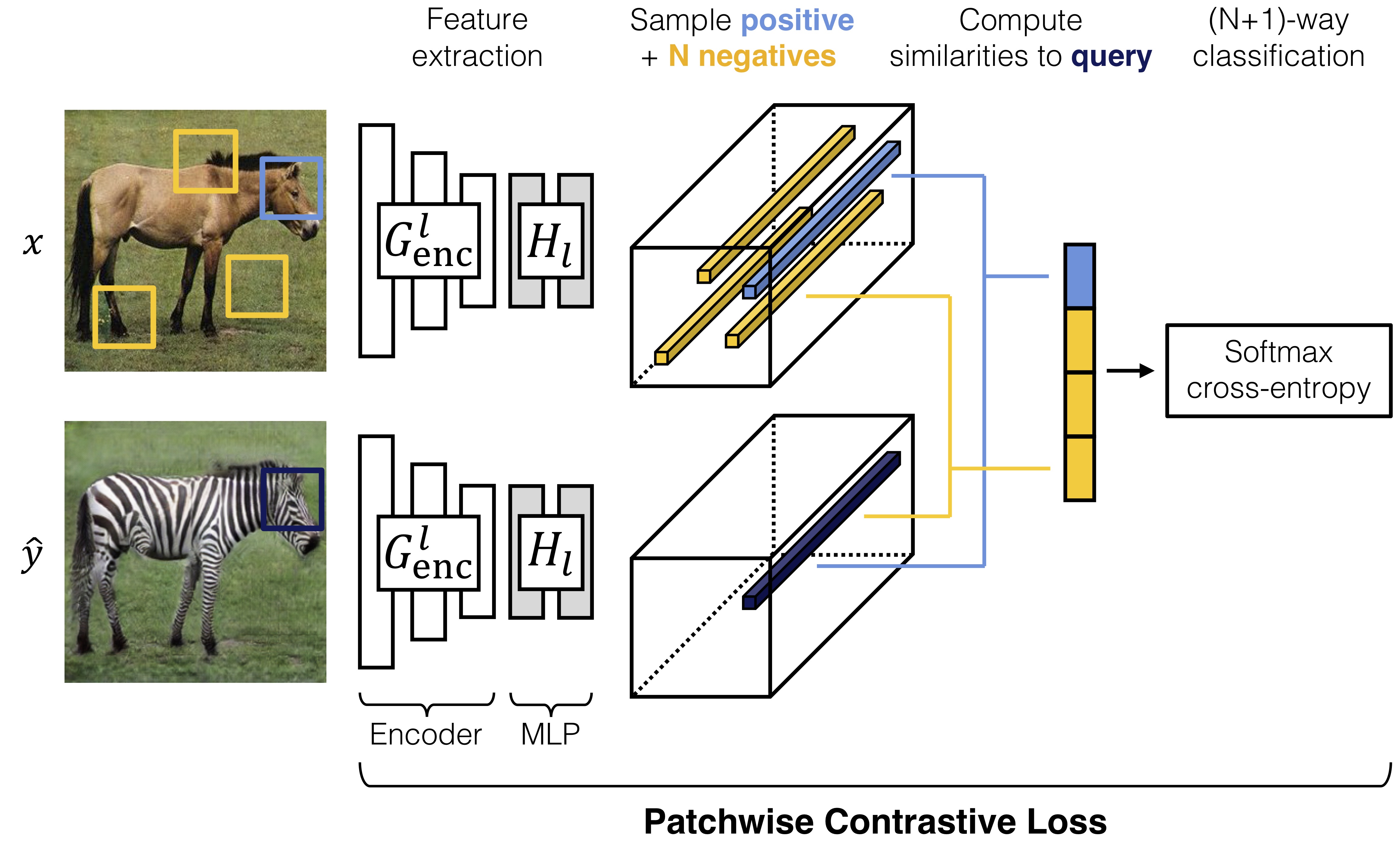}
\caption{A visual demonstration of the features extracted from an input image $\bm{x}$ and output image $\hat{\bm{y}}$ (Park et al., 2020). Corresponding patches are sampled from both images as well as $N$ other patches from the input image. These patches are used to calculate the PatchNCE loss in Eq.~\ref{patch_nce_loss} (Park et al., 2020). In the example presented here, high similarity between positive patches is desired to retain the shape of the animal head, while allowing re-coloring of the fur. Conversely, we should not expect to retain similarities between the head and other parts of the body. The image is taken from Park et al. (2020).}
\label{patch_nce}
\end{figure}

Contrastive unpaired translation is a GAN architecture that enables one-sided image-to-image translation (Park et al., 2020). In addition to the adversarial loss, which is dependent on the networks $G$ and $D$, CUT provides the PatchNCE loss and feature network $H$. The PatchNCE loss is used both to retain mutual information between the input image $\bm{x}$ and output $\hat{\bm{y}}$ as well as to enforce the identity translation $G(\bm{y}) = \bm{y}$ (Park et al., 2020). The feature network is defined as the first half of the generator, the encoder, plus a multi-layer patchwise (MLP) network with two layers and is used to encode the input and output images into feature tensors. Patches from the output image $\bm{\hat{y}}$ are sampled, passed to the feature network, and compared to the corresponding (positive) patch from the input as well as $N$ other (negative) patches from the input image. The process is shown in Figure~\ref{patch_nce} (Park et al., 2020).

Since $G_{enc}$ is used to translate a given image, its feature stack is readily available, with each layer and spatial position corresponding to a patch in the input image. We select $L$ layers from the feature map and pass each layer through the patchwise network to produce features $\{ \bm{z}_l \} = \{ H_l(G_{enc}^l (\bm{x})) \}_L$ where $G_{enc}^l$ is the output of the $l$-th layer. The feature $\bm{z}_l^s$ then represents the $s$-th spatial location of the $l$-th layer and $\bm{z}_l^{S \backslash s}$ represents all other locations. The PatchNCE loss is given by (Park et al., 2020)
\begin{equation}
\mathcal{L}_{PatchNCE} (G, H, X) := \mathbb{E}_{\bm{x} \sim X} \sum_{l=1}^{L} \sum_{s=1}^{S_l} \ell (\hat{\bm{z}}_l^s, \bm{z}_l^s, \bm{z}_l^{S \backslash s}),
\label{patch_nce_loss}
\end{equation}
where the sums are taken over all desired layers $l$ and spatial locations $s$ within each layer,
\begin{equation}
\ell(\bm{z}, \bm{z^+}, \bm{z^-}) := -\log{ \left[ \frac{\exp{(\bm{z} \cdot \bm{z^+} / \tau)}}{\exp{(\bm{z} \cdot \bm{z^+} / \tau)} + \sum \limits_{n=1}^N \exp{(\bm{z} \cdot \bm{z_n^-} / \tau)}} \right] },
\end{equation}
where $N$ is the number of negative samples and the temperature $\tau = 0.07$ scales the magnitude of penalties on the negative samples (Park et al., 2020). The overall loss used for network training is
\begin{equation}
\mathcal{L} (G, D, H, X, Y) = \mathcal{L}_{GAN} (G, D, X, Y) + \lambda_X \mathcal{L}_{PatchNCE} (G, H, X) + \lambda_Y \mathcal{L}_{PatchNCE} (G, H, Y),
\end{equation}
where the weighing factors $\lambda_X = \lambda_Y = 1$ when using the default CUT options (Park et al., 2020). The first PatchNCE loss term is used to retain mutual information between the input image $\bm{x}$ and output $\hat{\bm{y}}$. The second is used as an identity loss term, where translating the input image $\bm{y}$ should produce $\bm{y}$ as the output. \medbreak

The CUT training scripts provide several network definitions for the generator, discriminator, and feature network. For the work in this paper, we use the default options where the generator is a ResNet-based architecture that consists of 9 ResNet blocks between up and down-sampling layers and the discriminator is a 70x70 PatchGan network that originates from the work of Isola et al. (2016). \medbreak

The remainder of this section provides an overview of the datasets used to train several CUT generators.

\subsection{Target Domain -- Outdoor Image Dataset}
\label{outdoor_data}

\begin{figure}[H]
\centering
\includegraphics[width=\linewidth]{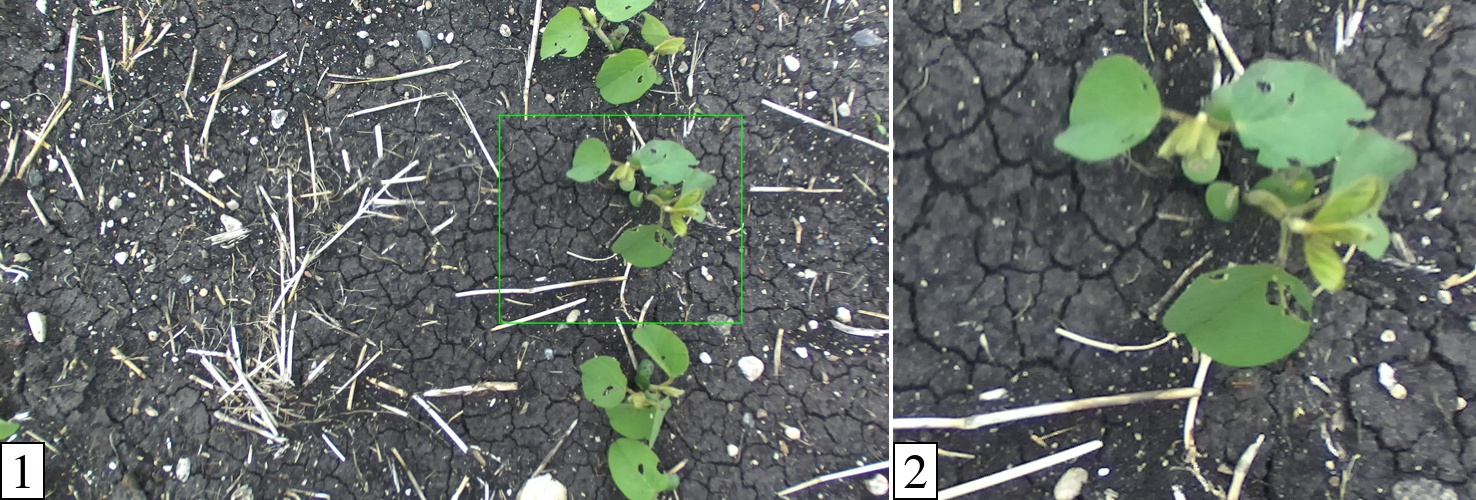} 
\caption{Outdoor multi-plant image of soybean (1) and cropped single-plant image (2). The cropping bounds, shown by the green bounding box, are determined by hand. The single-plant images constitute the target domain $Y$ for generator training.}
\label{example_outdoor_data}
\end{figure}

Generator training requires a dataset of outdoor images to represent the target domain $Y$ for image translation. We construct our outdoor dataset by sampling from the 540000 available images in the outdoor image database (Beck et al., 2021). The images are individual frames from videos taken with a camera mounted to a tractor while travelling through a field. The full-resolution images (2208x1242 px) often contain several plants with unknown locations and must be cropped by hand to obtain single-plant photos suitable for image translation. Manual cropping is a time-consuming process and, in general, limits our overall training dataset size. For initial experiments, we use 64 hand-cropped single-plant field images to construct the target dataset. We perform larger-scale experiments that include image translation of several species with 512 single-plant field images. Table~\ref{dataset_parameters} summarizes the relevant parameters for all training datasets. Example multi- and cropped single-plant outdoor images are shown in Figure~\ref{example_outdoor_data}.

The outdoor plant datasets used for model training contain several single-plant images of canola, oat, soybean, and wheat. The cropping bounds are varied to provide data with plants of several sizes relative to the image, this helps to prevent the generator from increasing or decreasing the size of the plant in the image during translation. Similarly, we ensure that plants appear in all locations of the single-plant images, such as centre, top-left, etc. to minimize positional drift. Additional images sampled from this dataset can be seen in Figures~[\ref{cropped_lab_images},~\ref{uncorrected},~\ref{soybean},~\ref{all}](1).

\subsection{Input Domain -- Cropped Lab Image Dataset}
\label{lab_data}

The first dataset to be used as the input domain $X$ to the generator consists of several indoor single-plant images with a blue screen background. These images are provided by the EAGL-I system, as described in (Beck et al., 2020). EAGL-I employs a GoPro Hero 7 camera mounted on a movable gantry capable of image capture from positions that vary in all three spatial dimensions. Additionally, the gantry includes a pan-tilt system to provide different imaging angles. For the purpose of this work, we attempt only to translate top-down images, thus we include images only where the camera is perpendicular to the floor, within a range of $\pm 10^\circ$. \medbreak

\begin{figure}[H]
\centering
\includegraphics[width=\linewidth]{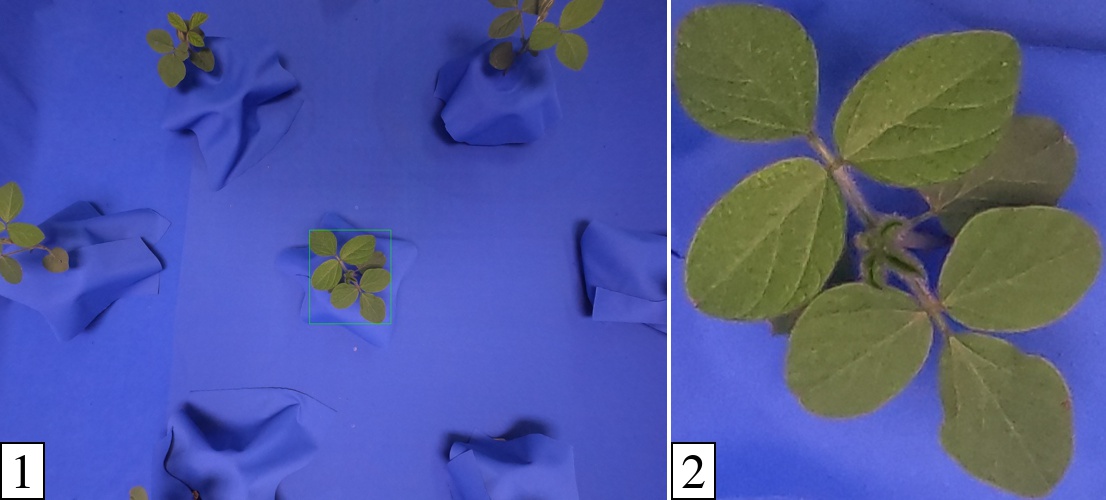}
\caption{Indoor multi- (1) and single-plant (2) images of soybean. The cropping bounds (shown by the green bounding box) are found automatically and do not require user input. The single-plant images form the input domain $X$ for generator training.}
\label{example_indoor_data}
\end{figure}

Similar to the previous section, the full-resolution images (4000x3000 px) contain several plants and must be cropped to obtain single-plant photos for training. However, in this case, both the plant and camera positions within the setting are known, and loose bounding boxes are found through geometric calculation. Tighter bounding boxes are obtained through an algorithm given in Appendix~\ref{michaels_bbox_code}, avoiding the need for hand-cropping. Figure~\ref{example_indoor_data} shows an example of both a multi- and single-plant lab image. \medbreak

To construct our dataset for generator input, we use 64 single-plant lab images as shown in Figure~\ref{example_indoor_data}(2). We expect that a translated image would not only contain a plant with characteristics that reflect the outdoor plant domain, but also contain soil that has been generated in place of the blue screen background.

\subsection{Input Domain -- Composite Image Dataset}
\label{composite_data}

As an alternative to the single-plant indoor photos of Section~\ref{lab_data}, we can use image processing techniques to remove the blue screen in a single-plant image and replace it with a real image of soil. These images, henceforth known as composite images, ideally require minimal translation to the image background and instead allow the generator to primarily translate plant characteristics. Such a network would be beneficial as it provides the user with the ability to influence the appearance of the background, even after passing through the generator. \medbreak

Soil backgrounds are randomly sampled from a set of images to provide sufficient variation in the training data. These backgrounds are hand-cropped from real outdoor data, so the soil is visually consistent with our target domain. Typically, we use 32 soil backgrounds for datasets composed of 64 composites. See Table~\ref{dataset_parameters} for the exact number of backgrounds used to construct each dataset. \medbreak

\begin{figure}[H]
\centering
\includegraphics[width=\linewidth]{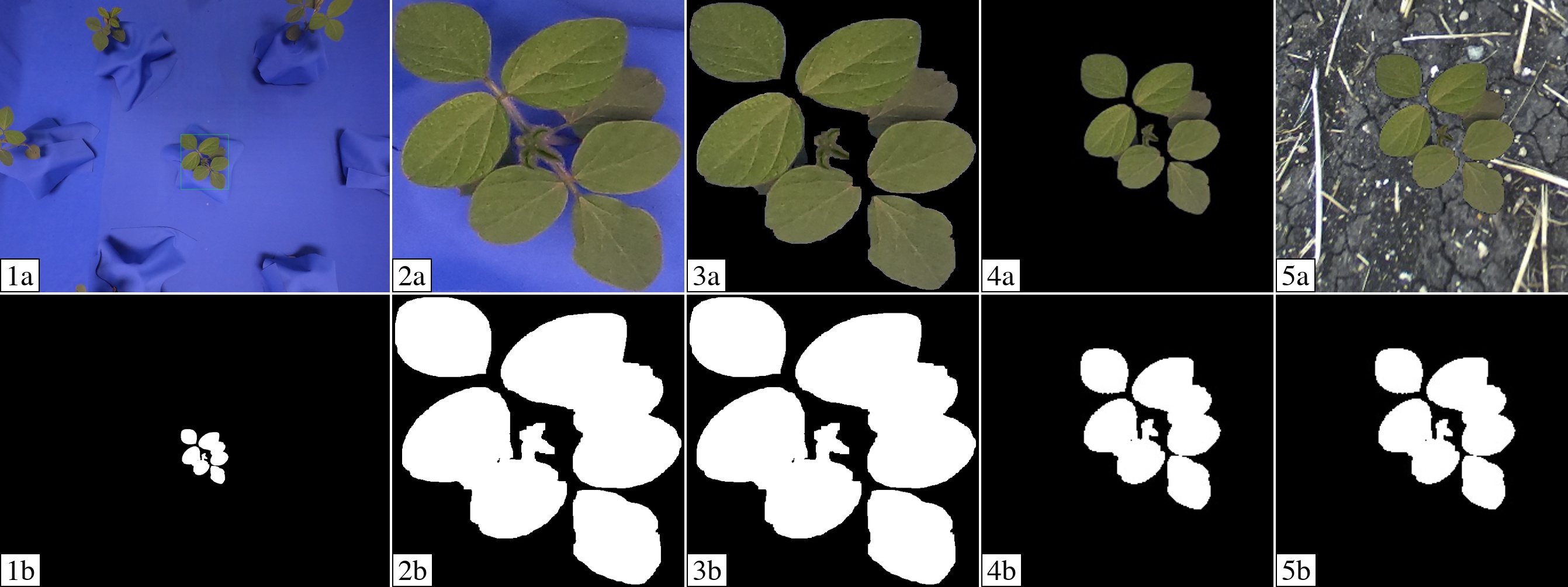} 
\caption{Intermediate images created during composite image formation (a) and their associated binary masks (b). The multi-plant image (1) is cropped to obtain a single-plant image (2). The single-plant binary mask is used to remove the blue screen in the single-plant image (3). The single-plant image is then resized and padded to give the effect of random placement within the final image (4). We join a soil background to the resized image to create the final composite (5). Images such as (5a) are used to form the input domain $X$ for generator training.}
\label{composite_data_steps}
\end{figure}

Given an indoor multi-plant image and soil background, the steps for composite formation are listed below. This process is fully-automated through scripts written in Python that utilize the OpenCV and NumPy libraries for image processing. Example images that depict each step are shown in Figure~\ref{composite_data_steps}.

\begin{enumerate}
\item{Mask and create bounding boxes using a multi-plant image from the indoor plant database. Blue screen usage makes background removal a trivial problem that can be solved by image thresholding (Beck et al., 2020).}
\item{Crop the multi-plant image and masked multi-plant image to create single-plant images.}
\item{Remove the blue screen in the single-plant image with the single-plant mask.}
\item{Randomly re-scale the single-plant image size relative to the final output size. We choose the minimum and maximum scale values to be $S_{min} = 0.50$ and $S_{max} = 0.85$, respectively. The image size required for the CUT generator is 256x256 px, so the plant is scaled relative to this size. Plant scale is calculated by dividing the longest side of the single-plant image by the output image size. For example, if the single-plant image has dimensions 150x200 px and the output image has dimensions 256x256 px, then the scale is $200/256 = 0.78$. Finally, we pad the resized image with black pixels so the final image has size 256x256 px. Random padding gives the effect of random placement of the plant within the composite.}
\item{Combine the resized plant image with a randomly selected soil background.}
\end{enumerate}

In many cases, differences in lighting between indoor and outdoor images leave little contrast between the plant and soil background in a composite image. In preliminary generator training experiments, insufficient contrast could lead to the generator translating plant leaves to appear as soil and constructing a fully-synthetic plant in a different region of the image. Such an occurrence is detrimental for creating synthetic object detection data as we lose the ability to accurately provide a bounding box for our plant. In the datasets described below, color-correction is used during composite formation to match the plant color with real field plants and increase the contrast between the plant and soil to allow the generator to maintain semantic information after domain translation. Color-correction techniques such as histogram matching were explored but were not used in favour of simple mean matching since we still expect the generator to adjust plant color. The chosen color-correction process is described as follows. \medbreak

\begin{figure}[H]
\centering
\includegraphics[width=\linewidth]{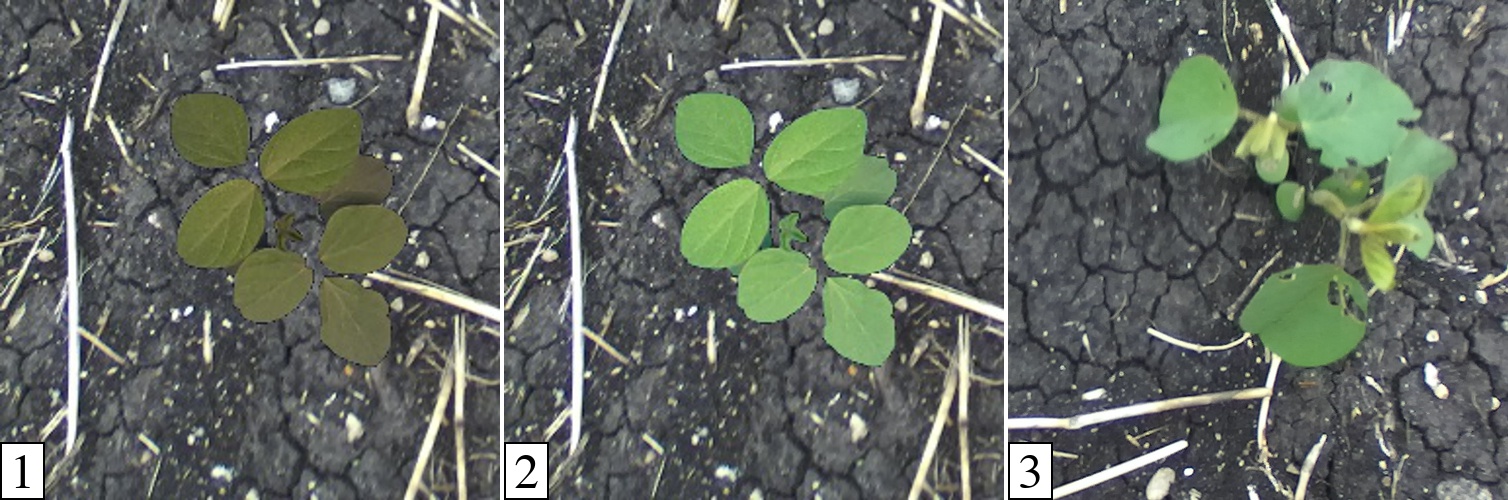}
\caption{Composite image of a soybean plant without (1) and with (2) color-correction. The color-corrected image contains plant leaves that better match the outdoor domain (3). We can use images such as (1) or (2) to provide the input domain $X$ for generator training.}
\label{color_corrected}
\end{figure}

Given an ($M \times N \times 3$) CIELAB single-plant image $\bm{x}$ and the associated ($M \times N$) binary mask $\bm{m}$, we want to correct the pixels belonging to the plant to have color that is consistent with the real outdoor plants in the domain $Y$ and leave all other pixels unchanged. To do this, we first determine the average value for the $k$-th channel of $\bm{x}$, only where the mask is true, given by
\begin{equation}
\bar{x}_k = \frac{ \sum\limits_{i=1}^M \sum\limits_{j=1}^N x_{ijk} m_{ij} }{ \sum\limits_{i=1}^M \sum\limits_{j=1}^N m_{ij} }.
\label{average}
\end{equation}
Note that Eq.~\ref{average} is simply the weighted average for all pixels of a given channel. The weight of each pixel is given by the value of the corresponding pixel in the binary mask, which have values 0 and 1 for false and true, respectively. We now construct our color-corrected image $\bm{x}^\prime$, where the individual pixels have values
\begin{equation}
x^\prime_{ijk} = x_{ijk} + m_{ij} (\bar{x}^\prime_k - \bar{x}_k)
\end{equation}
and $\bar{x}^\prime_k$ is the desired average value for the $k$-th channel of $\bm{x}^\prime$. We found that choosing the CIELAB channels to have average values $\bar{x}^\prime_L = 170$, $\bar{x}^\prime_A = 100$, and $\bar{x}^\prime_B = 160$ provides a suitable plant color. Note that in the definition of $x^\prime_{ijk}$, we can assign pixel values that exceed the valid range [0, 255]. In this case, we constrain $x^\prime_{ijk}$ to the same range. In general, color-correction leads to a composite image where the plant has enhanced lightness, greenness, and yellowness. An example of a composite image with and without color-correction is shown in Figure~\ref{color_corrected}.

For small-scale translation experiments, we construct our input domain to consist of 64 (optionally color-corrected) composite images. Larger-scale experiments use datasets composed of 512 composites. The method and parameters used to construct each dataset are summarized in Table~\ref{dataset_parameters}.

\section{Single-Plant Image Translation Results}
\label{results}

In this section we provide single-plant translation results for several GANs trained on the datasets described in Section~\ref{methods}. Training hyperparameters are consistent for all models, where we train for a total of 400 epochs and use a dynamic learning rate that begins to decay linearly to zero after the 200th epoch.

\begin{table}[H]
\footnotesize
\begin{center}
\begin{tabular}{l c c c c c c c c} \hline
Dataset Name & $N_x$ & $N_y$ & Species & Age (days) & $N_{\mathrm{backgrounds}}$ & $S_{\mathrm{min}}$ & $S_{\mathrm{max}}$ & Figure \\ \hline
Cropped Lab & 64 & 64 & Soybean & 10-40 & N/A & 1.00 & 1.00 & \ref{cropped_lab_images} \\ \hline
Composites & 64 & 64 & Soybean & 10-40 & 32 & 0.50 & 0.85 & \ref{uncorrected} \\ \hline
Color-Corrected Composites 1 & 64 & 64 & Soybean & 10-40 & 32 & 0.50 & 0.85 & \ref{soybean} \\ \hline
Color-Corrected Composites 2 & 512 & 512 & All & Varies & 128 & 0.50 & 0.85 & \ref{all} \\ \hline
\end{tabular}
\end{center}
\caption{Construction parameters for each training dataset. For the \textit{Cropped Lab} dataset, background replacement is not applicable since composite images are not created. Furthermore, the plant always makes up the entire image, so the scale is always 1.00. The \textit{Color-Corrected Composites 2} dataset contains each of the four available species; canola, oat, soybean, and wheat. These plants have minimum-maximum age ranges of 10-40, 0-365, 10-40, and 0-365 days, respectively.}
\label{dataset_parameters}
\end{table}

\subsection{Cropped Lab Dataset}

\begin{figure}[H]
\centering
\includegraphics[width=\linewidth]{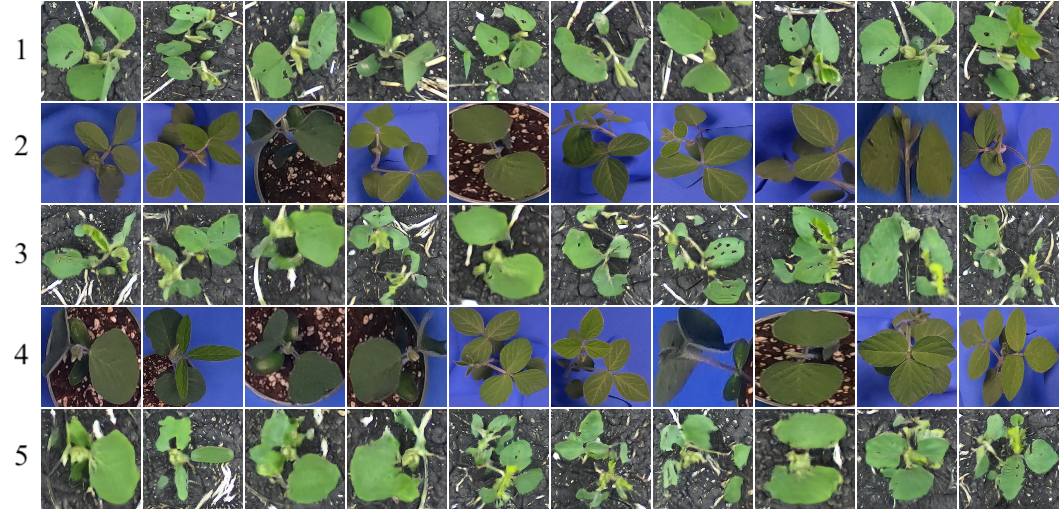}
\caption{Soybean images sampled from the distribution $Y$ (1), images sampled from the distribution $X$ (2), translated images $G(\bm{x})$ (3), images sampled from the distribution $V$ (4), and translated images $G(\bm{v})$ (5).}
\label{cropped_lab_images}
\end{figure}

Training a generator on the \textit{Cropped Lab} dataset allows us to directly translate indoor plants to outdoor, without the intermediate step of creating composite images. This dataset contains 64 indoor and outdoor images of soybean (128 total), aged 10-40 days. 20 additional indoor photos of soybean from the same age range unseen during the training process compose the distribution $V$ for qualitative evaluation of the model. Figure~\ref{cropped_lab_images} shows translation results for this generator.

Referring to both the training and evaluation images, the generator appears to be capable of image translation between these two domains with little error. Details such as leaf color, leaf destruction, and soil are suitably added in each translated image. Leaf veins, which are greatly pronounced in some indoor images, are removed in the translated images, reflective of the images from the outdoor domain. Despite the ability to translate indoor plants to appear as true outdoor plants, the greatest drawback to a generator trained on these images is the inability to control the image background in the translated images. As discussed in Section~\ref{multi_plant_synthesis}, this is problematic for the construction of images with multiple plants, where we want consistency in the background throughout the image.

\subsection{Composites Dataset}

\begin{figure}[H]
\centering
\includegraphics[width=\linewidth]{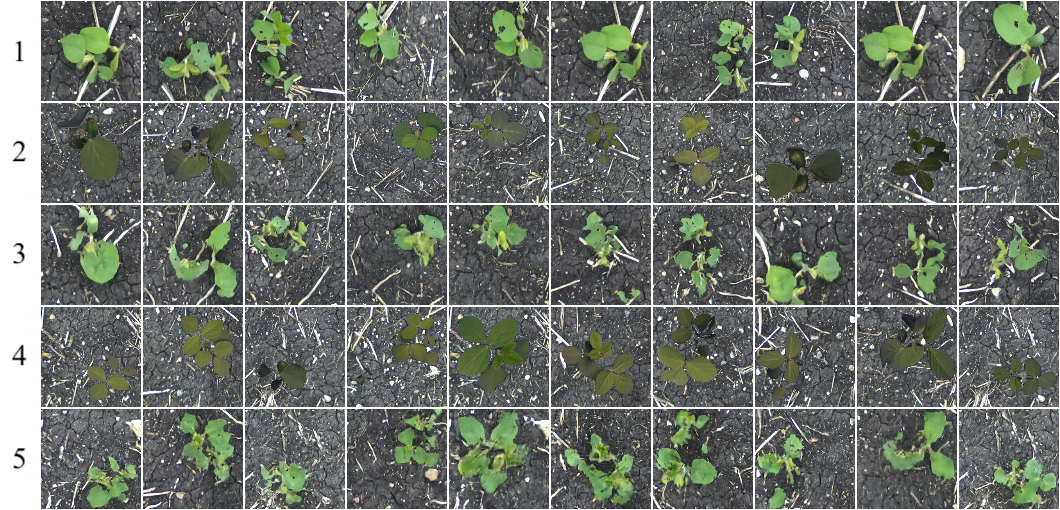}
\caption{Soybean images sampled from the distribution $Y$ (1), images sampled from the distribution $X$ (2), translated images $G(\bm{x})$ (3), images sampled from the distribution $V$ (4), and translated images $G(\bm{v})$ (5).}
\label{uncorrected}
\end{figure}

Training a generator on the \textit{Composites} dataset allows us to translate uncorrected composite images to outdoor-appearing plants. Here, we expect that the background in the output image remains consistent with the input. This dataset contains 64 composite and outdoor images of soybean (128 total), aged 10-40 days. 20 additional composite photos of soybean from the same age range unseen during the training process compose the distribution $V$ for qualitative evaluation of the model. Figure~\ref{uncorrected} shows translation results for this generator.

Referring to both the training and evaluation images, the generator is capable of image translation between these two domains. Similar to the previous section, plant details such as leaf color, destruction, and veins are suitably added/removed in the translated images. In general, the image backgrounds are consistent before and after translation. Small changes, such as the addition/removal of straw or pebbles in the background, are seen in the translated images. However, these details are sufficiently small so as not to have any negative effect during the construction of a synthetic multi-plant image.

\subsection{Color-Corrected Composite Datasets}

\begin{figure}[H]
\centering
\includegraphics[width=\linewidth]{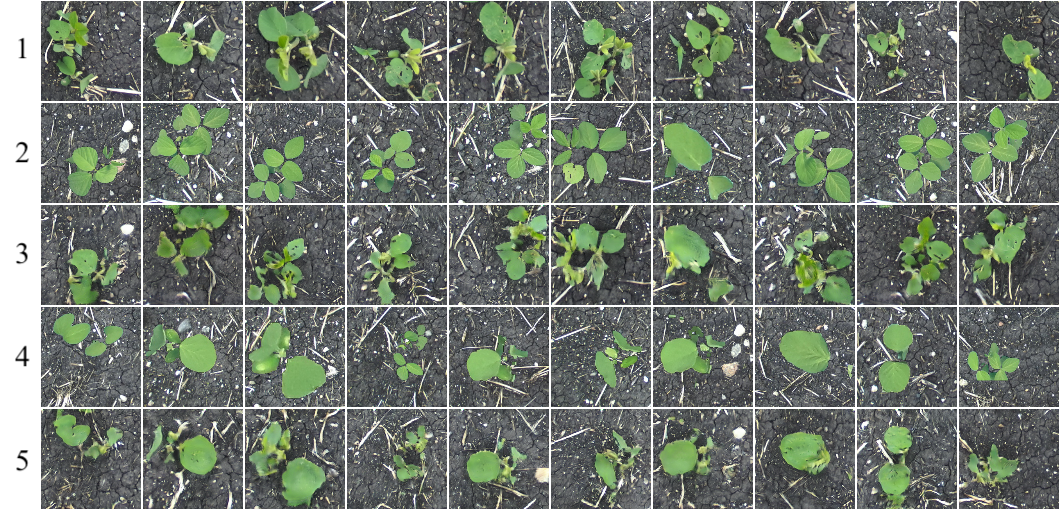}
\caption{Soybean images sampled from the distribution $Y$ (1), images sampled from the distribution $X$ (2), translated images $G(\bm{x})$ (3), images sampled from the distribution $V$ (4), and translated images $G(\bm{v})$ (5).}
\label{soybean}
\end{figure}

Training a generator on the \textit{Color-Corrected Composite} datasets allows us to translate color-corrected composite images to outdoor-appearing plants. Similar to the previous section, we expect that the background in the output image remains consistent with the input. The first generator in this section is trained on a dataset that contains 64 color-corrected composite and outdoor images of soybean (128 total), aged 10-40 days. 20 additional color-corrected composite photos of soybean from the same age range unseen during the training process compose the distribution $V$ for qualitative evaluation of the model. Figure~\ref{soybean} shows translation results for this generator. Additional translation results for generators trained on similar datasets of different species are shown in Appendix~\ref{additional_results}.

Referring to both the training and evaluation images, the generator is capable of image translation between these two domains. Similar to the previous section, plant details are suitably added or removed in the translated images, and the image background is consistent before and after translation. Additionally, we see little-to-no positional drift between plants in the input and output images, indicating a model such as this could be used for multiple-plant image synthesis. \medbreak

\begin{figure}[H]
\centering
\includegraphics[width=\linewidth]{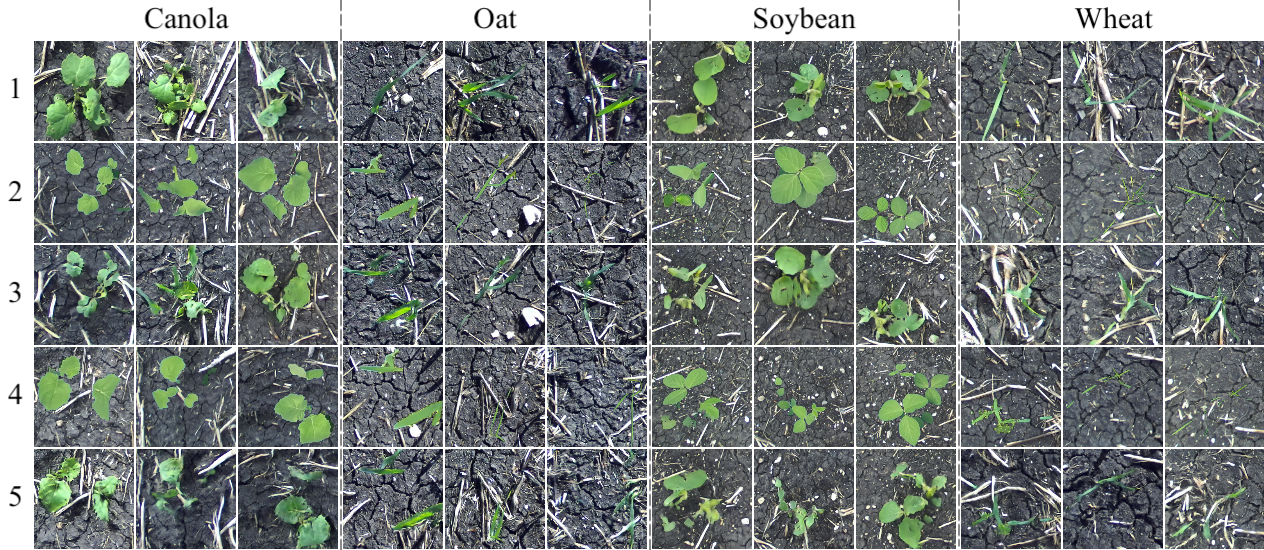}
\caption{Field images sampled from the distribution $Y$ (1), images sampled from the distribution $X$ (2), translated images $G(\bm{x})$ (3), images sampled from the distribution $V$ (4), and translated images $G(\bm{v})$ (5).}
\label{all}
\end{figure}

We now turn our focus to translating images of additional plant species. The \textit{Color-Corrected Composites 2} training dataset contains 128 color-corrected composite and 128 field images of each of the four available species; canola, oat, soybean, and wheat (1024 images total). An additional 20 color-corrected composites for each species compose the distribution $V$ for qualitative evaluation of the model. Image translation results are given in Figure~\ref{all}.

From the above translation results, we see that our method is easily extendable to additional species. Translated plants have no noticeable positional drift and background appearance is consistent between the input and output. A model such as the one providing image translations in Figure~\ref{all} could be used to construct synthetic multiple-plant images in a field setting with multiple plant species while retaining the original bounding boxes to enable the production of object detection data.

\section{Multiple-Plant Image Translation}
\label{multi_plant_synthesis}

\begin{figure}[H]
\centering
\includegraphics[width=\linewidth]{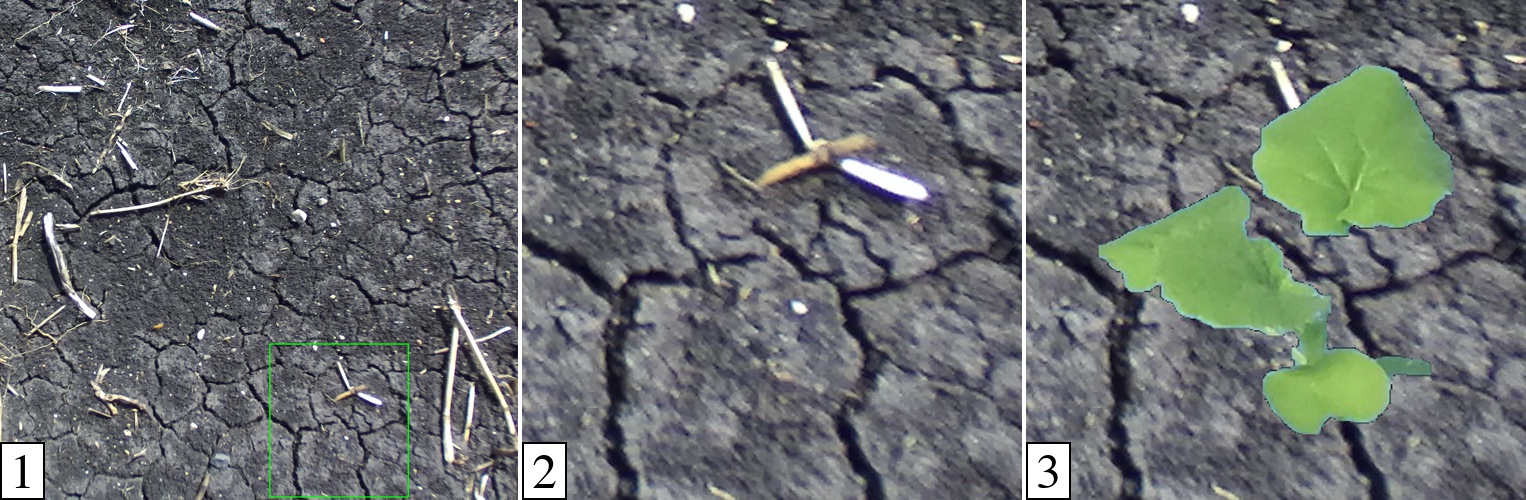} 
\caption{Full-scale soil image (1), cropped section of the image (2), and the color-corrected composite image to be passed to the generator (3). The cropping bounds are given by the green bounding box in (1).}
\label{background_selection_example}
\end{figure}

The results in the previous section suggest that a generator trained to translate composite images would be suitable for multi-plant image construction. With a single-plant translation generator, we can construct multi-plant images through the following algorithm. The algorithm makes use of existing full-scale soil images (see Figure~\ref{background_selection_example}(1) for an example) and indoor single-plant images with their binary masks (see Figures~\ref{composite_data_steps}(2a) and~\ref{composite_data_steps}(2b) for examples).

\begin{algorithm}[H]
\begin{algorithmic}[1]
\caption{Algorithm for constructing a synthetic outdoor multiple-plant image}
\label{multi_synthetic_algorithm}
\State randomly select a full-scale soil image
\For{number of plants}
	\State randomly select an indoor plant image 
	\State select corresponding binary mask
	\State select a sub-section of the soil image as the composite background
	\State form a composite image using the image, mask, and background section
	\If{color-correct is true}
		\State color-correct the composite image
	\EndIf
	\State translate composite image
	\State color-correct the translated image
	\State replace background sub-section with translated composite image
\EndFor
\end{algorithmic}
\end{algorithm}

In Algorithm~\ref{multi_synthetic_algorithm} we are effectively creating several composites, translating them, and placing them back into the full-scale image. Instead of selecting composite backgrounds from a pre-defined set, we randomly crop small sections from 20 larger images. Figure~\ref{background_selection_example} provides an example of composite background selection from a large-scale soil image.

To improve generator performance when translating images with these new backgrounds, we contruct an additional training dataset with 512 new color-corrected composite images and train a new CUT-GAN. Here, the composite backgrounds are randomly selected sub-sections of the large-scale backgrounds, as opposed to randomly selecting from the set of 128 small backgrounds used in the dataset \textit{Color-Corrected Composites~2}. This training dataset uses the same 512 images as \textit{Color-Corrected Composites 2} for the target domain~$Y$. Table~\ref{new_dataset_parameters} summarizes the parameters for constructing our new dataset.

\begin{table}[H]
\footnotesize
\begin{center}
\begin{tabular}{l c c c c c c c} \hline
Dataset Name & $N_x$ & $N_y$ & Species & Age (days) & $N_{\mathrm{backgrounds}}$ & $S_{\mathrm{min}}$ & $S_{\mathrm{max}}$ \\ \hline
Color-Corrected Composites 3 & 512 & 512 & All & Varies & 512 & 0.50 & 0.85 \\ \hline
\end{tabular}
\end{center}
\caption{Parameters for the new training dataset. The dataset contains each of the four available species; canola, oat, soybean, and wheat. These plants are aged 10-40, 0-365, 10-40, and 0-365 days, respectively. Each composite image possesses a unique background that comes from randomly cropping small sections from 20 large soil images.}
\label{new_dataset_parameters}
\end{table}

In line 11 of Algorithm~\ref{multi_synthetic_algorithm}, we color-correct the composite image after translation. This is a necessary step to help ensure a consistent background of our multi-plant images after translation. Background correction becomes especially useful if we choose an image background that differs from our training data. If we were to train a model to detect plants in our synthetic images without correction, it is possible that the model learns to locate plants through inconsistencies in the background where the translated plant is placed. Background color-correction is used as an attempt to mitigate this risk. Additional processing can be done to improve the joining of the translated composite with the background, but is not shown in this article. \medbreak

\begin{figure}[H]
\centering
\includegraphics[width=\linewidth]{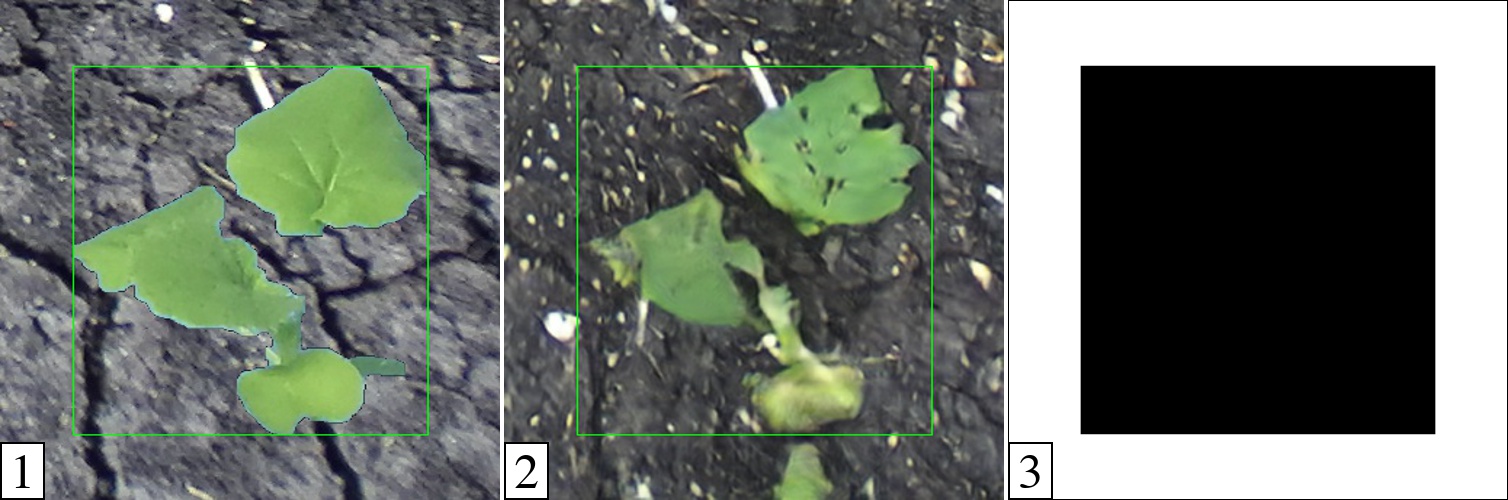} 
\caption{Color-corrected composite single-plant image (1), translated single-plant image (2), and the associated binary mask (3). The binary mask is false for all pixels within the bounding box and true for all other pixels.}
\label{background_correction_example}
\end{figure}

The procedure for background correction follows similarly to that of color-correction for composite images. However, rather than only correcting the pixels belonging to the plant, we correct all pixels in the translated image by adding a uniform offset to all pixels of a given channel. This offset is found by calculating the difference in average values for each RGB channel in the composite and translated images, only for pixels in a small region outside the plant bounding box. In general, for background correction, we assume that the translated plant does not exceed the bounding box. However, small patches of green outside the bounding box (see Figure~\ref{background_correction_example}(2)) lead to little difference in the result. \medbreak

We require three inputs: the ($M \times N \times 3$) composite single-plant image $\bm{x}$, ($M \times N \times 3$) translated single-plant image $\hat{\bm{y}}$, and the ($M \times N$) binary mask $\bm{m}$. Example images are given in Figure~\ref{background_correction_example}.

\begin{figure}[H]
\centering
\includegraphics[width=\linewidth]{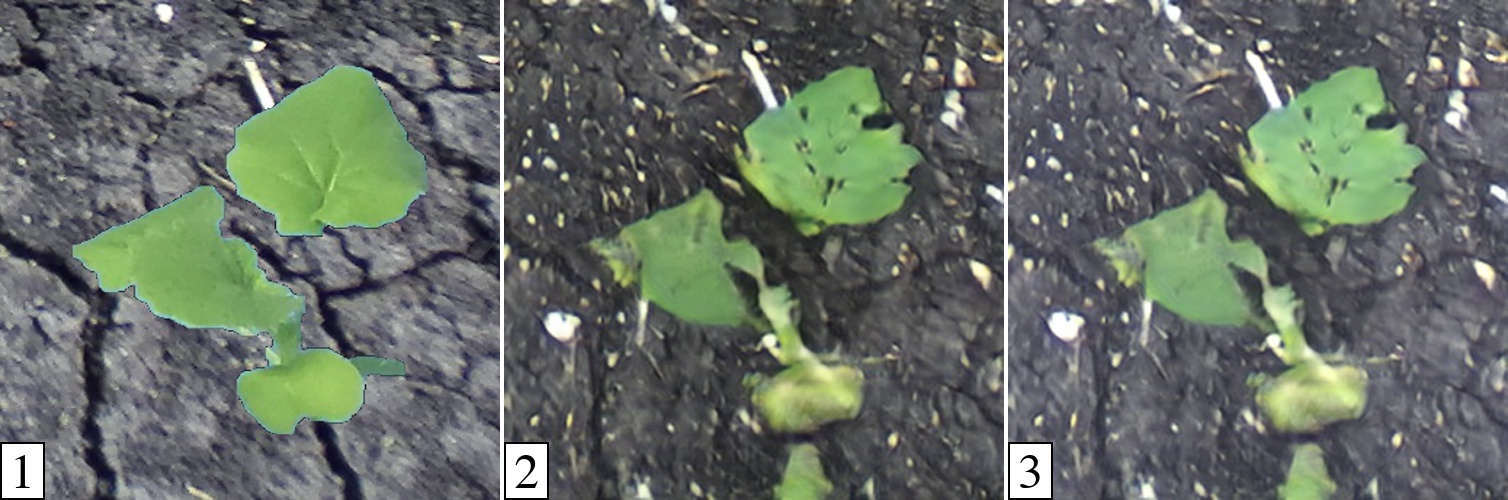} 
\caption{Single-plant images before (1) and after (2) translation and the corrected translated image (3). The soil color and lightness in (3) is more consistent with the input (1).}
\label{background_correction_single}
\end{figure}

\begin{figure}[H]
\centering
\includegraphics[width=\linewidth]{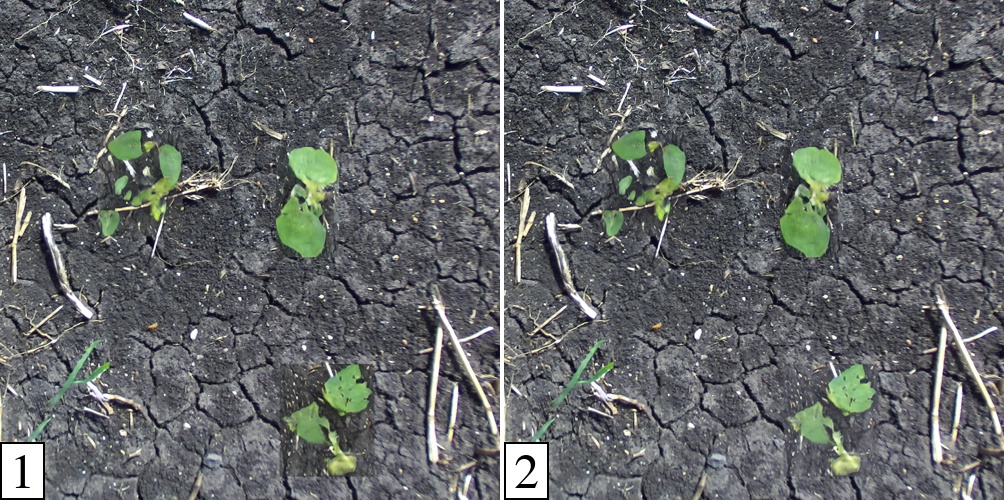} 
\caption{Translated multi-plant image without (1) and with (2) background correction. Note that the backgrounds of the translated sections of the image differ in color and lightness from the soil in the rest of the image. Background correction lessens this effect, especially seen for the soybean plant in the bottom-right.}
\label{background_correction_multi}
\end{figure}

We calculate the offset $d_k$ for the $k$-th channel of the corrected image $\hat{\bm{y}}^\prime$ as
\begin{equation}
d_k = \frac{ \sum\limits_{i=1}^M \sum\limits_{j=1}^N (x_{ijk} - \hat{y}_{ijk}) m_{ij} }{ \sum\limits_{i=1}^M \sum\limits_{j=1}^N m_{ij} }.
\label{background_correction}
\end{equation}
Equation~\ref{background_correction} can equivalently be considered as the difference in weighted averages for the input and output images where the weight of all pixels within the bounding boxes is zero and one elsewhere. The pixels of $\hat{\bm{y}}^\prime$ are then given by
\begin{equation}
\hat{y}^\prime_{ijk} = \hat{y}_{ijk} + d_k,
\end{equation}
constrained to the range $[0, 255]$. Figure~\ref{background_correction_single} shows the effect of translating a single-plant image with and without applying background correction afterwards. Figure~\ref{background_correction_multi} demonstrates the difference between multi-plant images with and without background correction.

\begin{figure}[H]
\centering
\includegraphics[width=\linewidth]{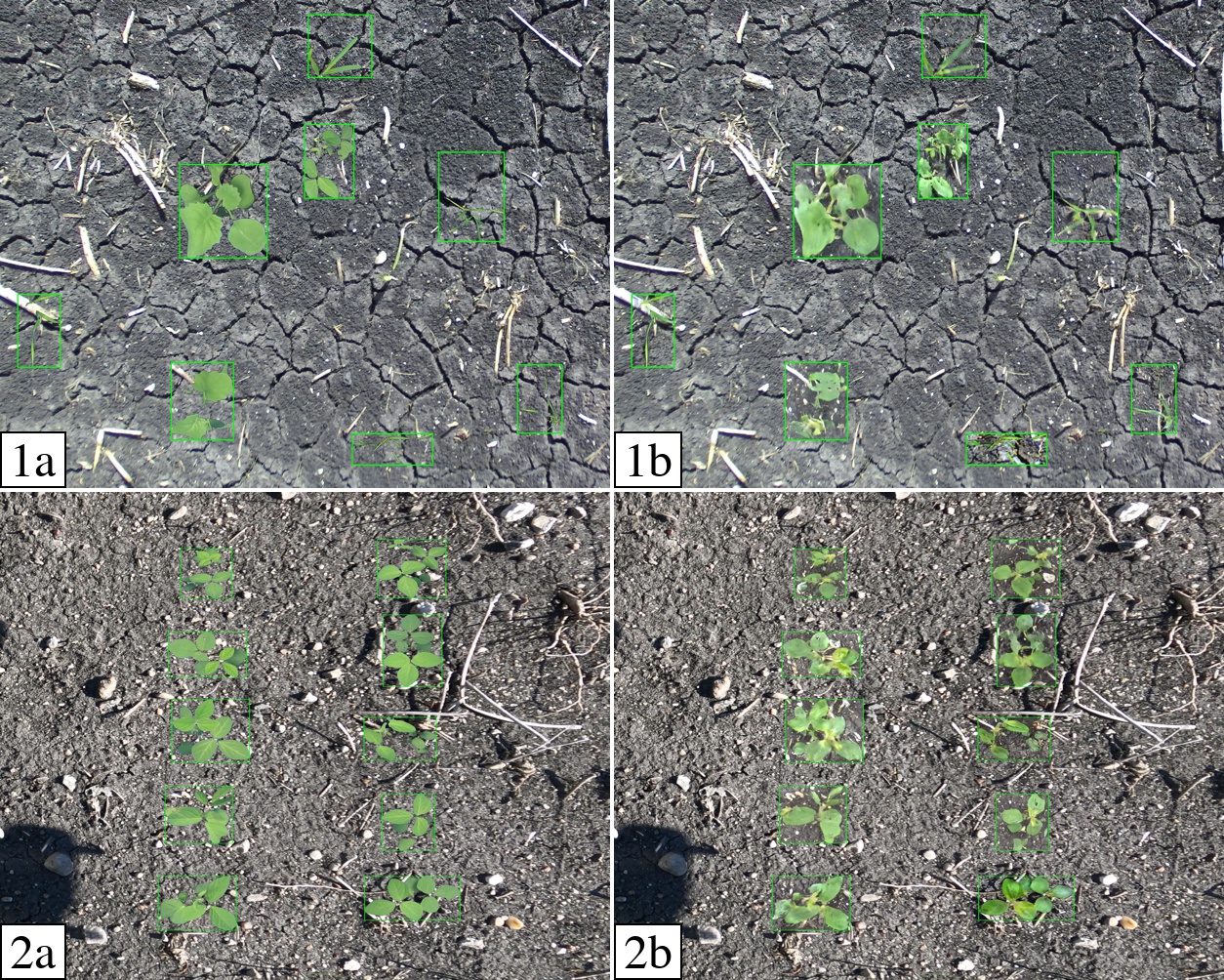}
\caption{Composite (a) and synthetic (b) multi-plant images where plants are placed randomly (1) or ordered into two rows (2). The randomly placed plants are chosen to be any of the four available species, the ordered plants are all soybean.}
\label{synthetic_multi_plant_examples}
\end{figure}

In general, the translation capabilities of the generator are invariant to the position of plants in a given image. As a result, there is no restriction on placement, aside from avoiding overlapping plants. We provide the option to place plants randomly, or by alignment into rows, as they would be in a production field. Additionally, the minimum and maximum scales of the plants relative to the background can be chosen by the user. Datasets can be made to contain plants of all species or individual plants can be selected. Examples of labeled synthetic multi-plant images are given in Figure~\ref{synthetic_multi_plant_examples}. From the multi-plant images presented here, we see that the position of each plant is maintained after translation, suggesting that such data could be useful for training a plant detection network.

\section{Plant Detection Using Augmented Datasets}
\label{plant_detection}

In general, object detection is a machine learning task that refers to the process of locating objects of interest within a particular image (Zhao et al., 2019). Models capable of object detection typically require training data consisting of a large number of images as well as the location and label of all objects within said images, usually given by bounding boxes. Such data is often produced by manual annotation, a procedure that is both time- and cost-intensive, especially as the number of objects within an image increases (Ayalew, 2020; Guillaumin and Ferrari, 2012).  As a result, object detection is an excellent test case of our image transformation methods.

As proof of concept that our synthetic multi-plant images can be beneficial for detection of real plants, we trained several YoloV5 (Bochkovskiy et al., 2020; Redmon et al., 2016) nano object detection models on various augmented datasets (Jocher et al., 2022). We choose the nano in particular as it has the fewest number of trainable parameters in comparison to all other YoloV5 networks. This is most desirable for our training datasets which contain few classes and little variability in the data itself. In all cases the network is trained to locate canola, oat, soybean, and wheat. However, we are not attempting to find a solution to a classification problem, so these four species are grouped into a single class named \textit{plant}. Three primary training dataset types are described below, and each includes the random placement of non-overlapping plants onto an image background. We use the size and location of the single plant image embedded within the background as the ground truth bounding boxes for our training data. \medbreak

\begin{figure}[H]
\centering
\includegraphics[width=\linewidth]{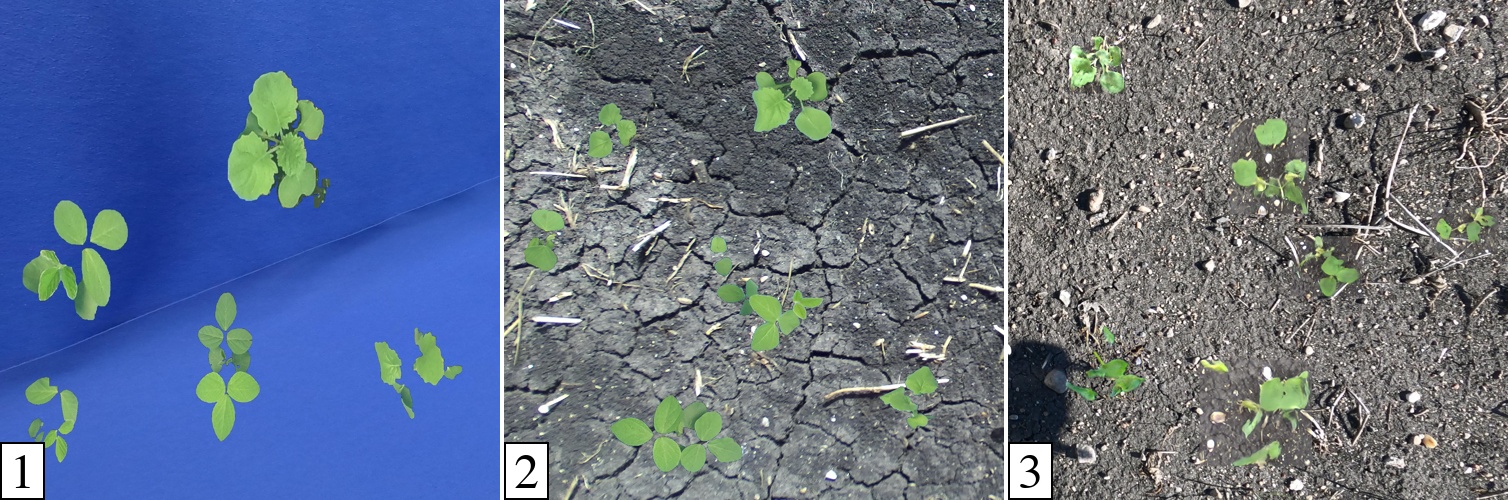}
\caption{Sample images from the \textit{Baseline} (1), \textit{Composite} (2), and \textit{GAN} (3) datasets used to train the YoloV5 nano networks. In all cases, plants are color-corrected and randomly placed into non-overlapping positions onto an image background. The \textit{GAN} images receive the additional step of plant translation.}
\label{yolo_training_images}
\end{figure}

The first training dataset contains 80000 blue screen images with color-corrected plants randomly placed throughout (see Fig~\ref{yolo_training_images}(1)). The dataset is split into training, validation, and testing sets with proportions 0.8, 0.1, and 0.1, respectively. Note that no background replacement or GAN is used, so the differences between the training data and real field data are significant. Hence, we expect the performance of this network to be poor in general. For future references this dataset is referred to as the \textit{Baseline} dataset. \medbreak

The \textit{Composite} dataset contains 80000 multi-plant color-corrected composite images (see Fig~\ref{yolo_training_images}(2)) with the same 0.8, 0.1, 0.1 split as above. Here we include a real soil background, but no GAN is used to individually translate each plant. Since this dataset is more similar to real data than the baseline we expect to see improved performance on an evaluation set composed of real images. \medbreak

The third dataset used to train the network consists of 80000 multi-plant GAN images (see Fig~\ref{yolo_training_images}(3)) split identically as above. This training data is created through the multi-plant GAN procedure previously described in this section. Here we expect network performance to be the greatest, since the training and target datasets are most similar. For future references this dataset is named \textit{GAN}.

An additional dataset, known as the \textit{Merged} dataset, is composed as the union of both the \textit{Composite} and \textit{GAN} datasets. As such, this training dataset consists of 160000 total images for the network with half of the images including the usage of a GAN for plant translation. \medbreak

\begin{figure}[H]
\centering
\includegraphics[width=\linewidth]{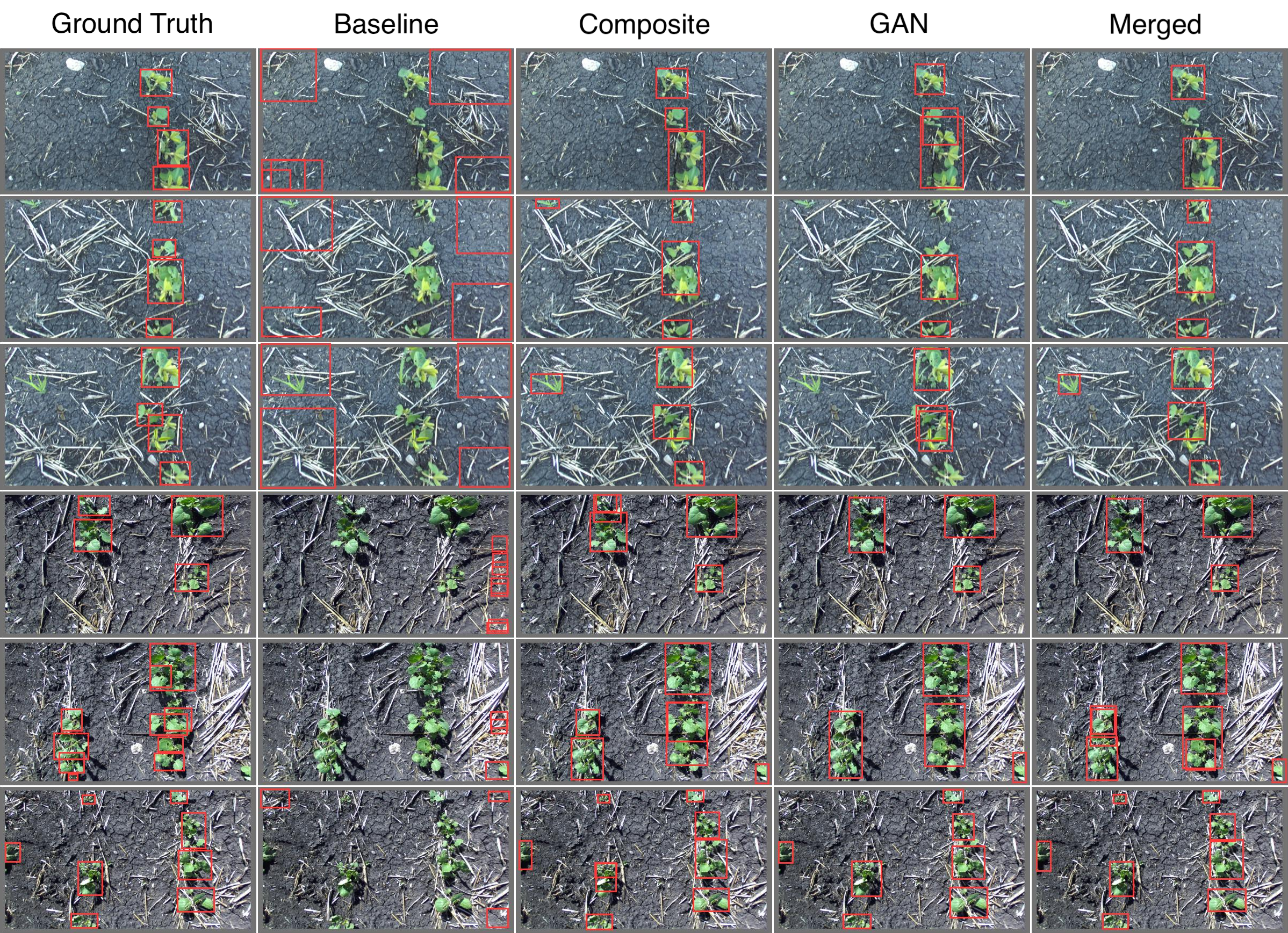}
\caption{Sample images labeled by YoloV5 nano object detection models trained on various augmented datasets (see labels). Note that the ground truth bounding boxes are determined by hand. The top three image rows contain only soybean plants, the bottom three contain only canola.}
\label{yolo_eval}
\end{figure}

With several plant detection models trained on varying datasets, we evaluate our models on 253 additional real multi-plant images of canola and soy. These images were excluded during the training process and ground truth bounding boxes are determined by hand. The evaluation metrics used are precision, recall, and mean average precision (mAP). Before defining the metrics, we must first consider the intersection over union (IoU), given by (Everingham et al., 2010)
\begin{equation}
\mathrm{IoU} := \frac{\mathrm{Area~of~Overlap}}{\mathrm{Area~of~Union}},
\end{equation}
where Area of Overlap refers to the area of overlap between the ground truth and predicted bounding boxes and Area of Union is the total area from joining the bounding boxes. Whether a bounding box prediction is considered to be successful is dependent on the IoU threshold, where we consider any prediction leading to an IoU greater than the threshold to be correct. Now, with IoU being used to determine if a model prediction is correct, the metrics precision and recall are defined by (Everingham et al., 2010)
\begin{align}
\mathrm{precision} &:= \frac{tp}{tp + fp}, \\ 
\mathrm{recall} &:= \frac{tp}{tp + fn}
\end{align}
where $tp$ denotes the number of true-positives, $fp$ the false-positives, and $fn$ the false-negatives. We interpolate precision over 101 recall values in the range [0.00, 1.00] in steps of 0.01. For notational simplicity, we define the set of recall values $R = \{ 0.00, 0.01, ..., 1.00 \}$. The interpolated precision is given by (Everingham et al., 2010)
\begin{equation}
\mathrm{precision}_{\mathrm{inter}}(r) := max\{ \mathrm{precision}_{\tilde{r}:\tilde{r} \geq r}(\tilde{r}) \}.
\end{equation}
Finally, mAP is defined by (Everingham et al., 2010)
\begin{equation}
\textrm{mAP} := \frac{1}{n} \sum_{k=1}^{n} \frac{1}{|R|} \sum_{r \in R} \mathrm{precision}_{\mathrm{inter}}(r),
\end{equation}
where $n$ denotes the number of classes and the outer sum leads to a mean precision for all classes. Note that since we group our species into a single class we have $n = 1$ and the outer sum is presented here only for verbosity. The notation mAP@0.5 denotes the evaluation of mAP using an IoU threshold of 0.5, consistent with the Pascal VOC evaluation metric (Everingham et al., 2010). Alternatively, mAP@0.5:0.95 denotes the average mAP value for all IoU thresholds in the range [0.50, 0.95], in steps of 0.05, which is identical to the evaluation metric for the COCO dataset challenge (Lin et al., 2014). Sampled visual results for the performance of each network are given in Figure~\ref{yolo_eval}, numerical results are provided by Table~\ref{yolo_metrics}, including both the mAP@0.5 and mAP@0.5:0.95 metrics.

\begin{table}[H]
\footnotesize
\begin{center}
\begin{tabular}{l c c c c} \hline
Training Dataset & precision & recall & mAP@0.5 & mAP@0.5:0.95 \\ \hline
\textit{Baseline} & 0.030 & 0.099 & 0.010 & 0.002 \\ \hline
\textit{Composite} & 0.597 & 0.468 & 0.454 & 0.202 \\ \hline
\textit{GAN} & 0.647 & 0.459 & 0.479 & 0.209 \\ \hline
\textit{Merged} & 0.673 & 0.500 & 0.528 & 0.240 \\ \hline
\end{tabular}
\end{center}
\caption{precision, recall, and mAP metrics when evaluating our four YoloV5 nano models on 253 real images of canola and soybean. The training dataset denotes the images used to train the network for which we are evaluating, not the set from which the evaluation images are taken.}
\label{yolo_metrics}
\end{table}

From both Figure~\ref{yolo_eval} and Table~\ref{yolo_metrics} it is clear that the baseline model shows no capacity to detect plants. This is expected as the training data is vastly different from the real images through which we evaluate the models. Both the \textit{Composite}- and \textit{GAN}-trained models show good ability to locate plants in the evaluation images, however the \textit{GAN}-trained model performs better according to both the mAP@0.5, and mAP@0.5:0.95 metrics. The combination of the \textit{Composite} and \textit{GAN} datasets led to the greatest performing network in terms of our metrics. However, this could come as a result of being exposed to twice the number of training images. In general, we see that the Yolo models trained on the \textit{Composite}, \textit{GAN}, and \textit{Merged} datasets all appear capable of plant detection on the provided images.

\section{Conclusion}
\label{conclusion}

The contribution of this work is an image translation process through which we can produce artificial images in field settings, using images of plants taken in an indoor lab. The method is easily extendable to new plant species and settings, provided there exists sufficient real data to train the underlaying GAN. The construction of our own augmented datasets enables further training of neural networks, with applications to in-field plant detection and classification. More importantly, our work is a first step in demonstrating that plants grown in growth chambers under precise and fully-controlled conditions can be used to easily generate large amounts of labelled data for developing machine learning models that operate in outdoor environments. This work has the potential to significantly improve and accelerate the model development process for machine learning applications in agriculture. 

Future work will consist of mimicking outdoor lighting conditions via controllable led-based growth-chamber lights, leading to more variety in the input data and hopefully more realistic synthetic data. Additional work will focus on using the approach presented here to develop in-field plant-classification models, as well as developing outdoor datasets and machine learning models for other problem domains (such as disease detection). This will include an investigation into whether classification tasks require their own species-specific GAN for the image translation process. 


\clearpage

\appendix

\section{Improved Indoor Bounding Box Algorithm}
\label{michaels_bbox_code}

Here we present an algorithm used to obtain tighter bounding boxes for our multi-plant lab images. The algorithm functions by mapping all sub-regions of a multi-plant image to the correct plants. For the purpose of this paper, we will refer to these regions as fragments. \medbreak

\begin{figure}[H]
\centering
\includegraphics[width=0.75\linewidth]{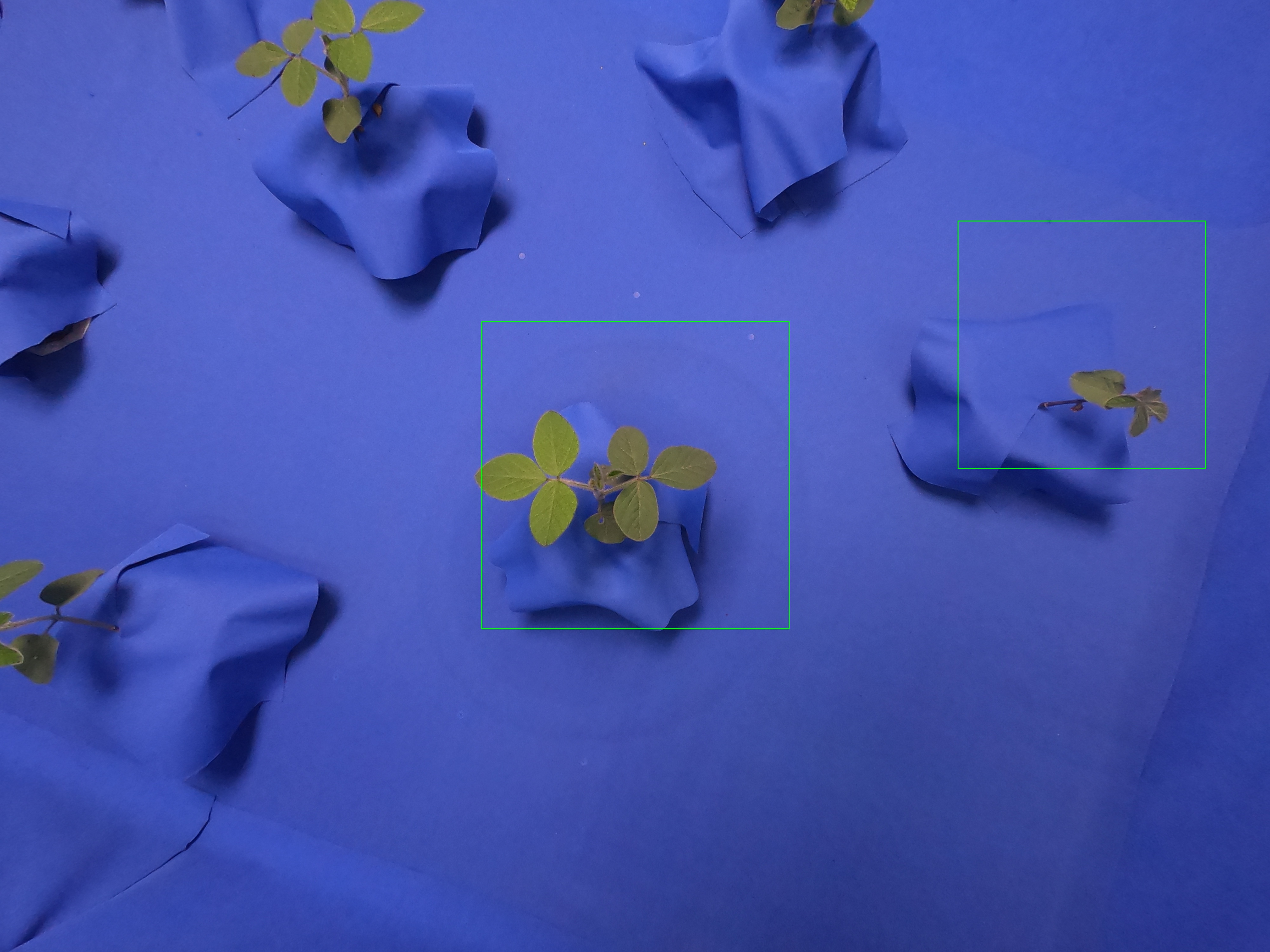}
\caption{Indoor multi-plant image with the original bounding boxes shown in green. A conservative estimate of the plant position leads to loose bounding boxes for each plant. Plants that are overlapping or are too close to the image border do not receive bounding boxes.}
\label{multi_bbox_initial}
\end{figure}

Initial, loose bounding boxes are immediately determined through geometric calculations relating the plant position in the scene and the camera position/angle. These initial bounding boxes include an additional tolerance to help ensure that the plants are fully contained within the box, resulting in a conservative estimate for the plant position within a given image. The loose bounding boxes are readily available in the indoor plant database when downloading the multi-plant images. An example of an indoor multi-plant image with the original bounding boxes is given by Figure~\ref{multi_bbox_initial}.

The first step is to remove the image background, here we convert a BGR image to CIELAB color space. The benefit to operating in CIELAB space is in the a- and b-channels of the image. The a-channel has a low magnitude for green pixels, and large for red. Similarly, the b-channel has a low magnitude for blue pixels, and large for yellow. We create an overall image mask through addition of a- and b-channel masks, where we set any pixel with a-value greater than \verb|a_cutoff| to zero and any pixel with b-value less than \verb|b_cutoff| to zero. The addition of the two masks results in a value of 0 anywhere the a-channel of a pixel is greater than \verb|a_cutoff| and the b-channel is less than \verb|b_cutoff| and a value of 255 for all other pixels in the image. In general, we assign the values \verb|a_cutoff| $= 130$ and \verb|b_cutoff| $= 95$. A border around the image is also constructed so that plants that reach the outside of the image are correctly labeled as fragments in the proceeding step. Algorithm~\ref{background_removal} provides the code, written in NumPy and OpenCV, used to construct the image mask.

\begin{algorithm}[H]
\caption{Algorithm for removing the multi-plant image background}
\begin{algorithmic}[1]
\State import cv2
\State import numpy as np
\State
\State \# Input: master = original image read as 3-dimensional array
\State
\State \# convert master into LAB color space and extract b-channel, a-channel
\State lab = cv2.cvtColor(master, cv2.COLOR\_BGR2LAB)
\State a\_channel = np.array(lab[ :, :, 1])
\State b\_channel = np.array(lab[ :, :, 2])
\State
\State \# masking
\State \_, b\_mask = cv2.threshold(b\_channel, b\_cutoff, 128, cv2.THRESH\_BINARY)
\State \_, a\_mask = cv2.threshold(a\_channel, a\_cutoff, 127, cv2.THRESH\_BINARY\_INV)
\State mask = (a\_mask + b\_mask)
\State mask[mask \textless~129] = 0
\State
\State \# introduce 1px frame around interior
\State mask[:, 0] = 0
\State mask[:, mask.shape[1] - 1] = 0
\State mask[0, :] = 0
\State mask[mask.shape[0] - 1, :] = 0
\end{algorithmic}
\label{background_removal}
\end{algorithm}

Next, we find all fragments in the masked image. Image opening (equivalent to erosion, followed by dilation) is used to remove small fragments in the image that would otherwise be difficult to assign to or may not belong to any plant. Then, we remove any fragments with size below \verb|size_cutoff|, typically we have \verb|size_cutoff| $= 200$ px. Algorithm~\ref{fragments_from_mask} provides the code for this operation.

\begin{algorithm}[H]
\caption{Algorithm for generating fragments from mask}
\begin{algorithmic}[1]
\State import cv2
\State import numpy as np
\State from scipy.ndimage import measurements
\State
\State \# open mask for fast removal of small fragments
\State kernel = np.ones((k\_size, k\_size), np.uint8)
\State thresholded = cv2.morphologyEx(mask, cv2.MORPH\_OPEN, kernel)
\State
\State \# label fragments
\State labels, n\_fragments = measurements.label(thresholded)
\State thresholded\_copy = thresholded.copy()
\State
\State \# remove small fragments, 0 is the background
\State for frag\_number in range(1, n\_fragments+1):
\State ~~~~if measurements.sum(thresholded, labels, frag\_number) \textless ~size\_cutoff*255:
\State ~~~~~~~~slices = find\_objects(labels == frag\_number)[0]
\State ~~~~~~~~thresholded\_copy[slices[0].start: slices[0].stop, slices[1].start: slices[1].stop] = 0
\State
\State \# relabel fragments
\State labels, n\_fragments = measurements.label(thresholded\_copy)
\end{algorithmic}
\label{fragments_from_mask}
\end{algorithm}

We now possess several image fragments with a unique label and must associate each to a plant in the original image. For each fragment $f \in F$, we calculate the centre of mass $c(f) = (x_f, y_f)$ where $x_f$ and $y_f$ are the x- and y-coordinates of the fragment, respectively. Additionally, for each plant $p \in P$, we find the centre of the plant $c(p) = (x_p, y_p)$ using the image metadata. We initialize a distance matrix $D \in \rm{Mat} (|P| \times |F|)$ where $D := (d(p, f))$ and $d(p, f)$ is the Euclidean distance between the centre of the plant $p$ and the centre of mass of the fragment $f$. \medbreak

\begin{figure}[H]
\centering
\includegraphics[width=0.75\linewidth]{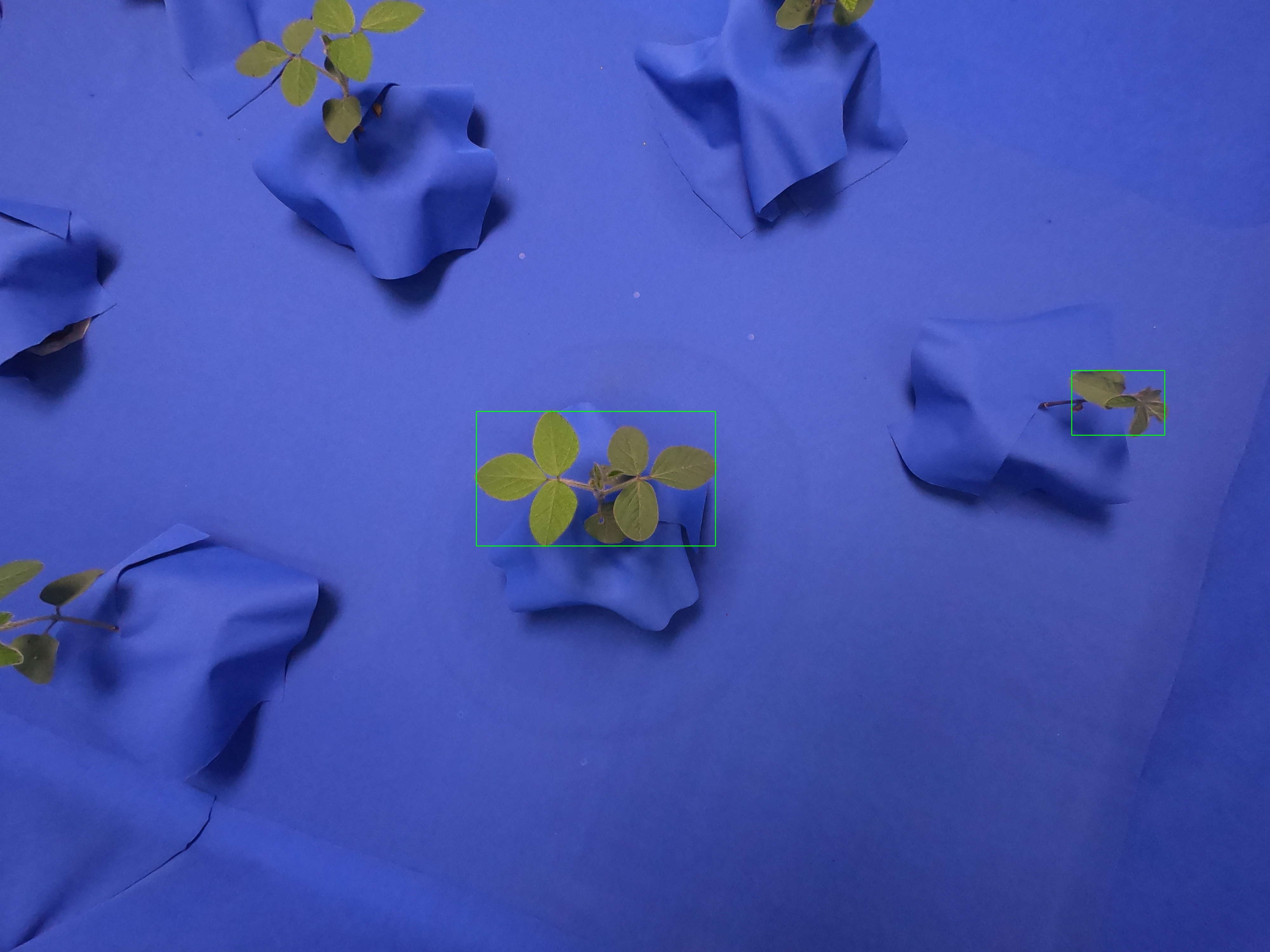}
\caption{Indoor multi-plant image with the improved bounding boxes shown in green. The bounding boxes are significantly tighter than the originals.}
\label{multi_bbox_final}
\end{figure}

Finally, we reduce extraneous fragments by finding any with an unreasonable distance to a plant in the original image. We estimate the radius of the plant in the image $r(p)$, this is done by calculating the width of the bounding box from the original metadata. Equivalently, the bounding box height could be used to calculate the radius since the original bounding boxes are all square. We remove all fragments that satisfy $D_{p, f} > r(p) \cdot T_1$ where $T_1$ is a distance threshold multiplier, typically assigned the value $T_1 = 1.6$. From the remaining valid fragments, we assign the fragment to the plant with the smallest Euclidean distance. The output of Figure~\ref{multi_bbox_initial} after the bounding box tightening algorithm is given in Figure~\ref{multi_bbox_final}.

\clearpage

\section{Additional Color-Corrected Composite Translation Results}
\label{additional_results}

This section contains image translation results for generators trained to translate color-corrected composite images of canola, oat, and wheat. The datasets used for generator training are similar to the \textit{Color-Corrected Composites 1} dataset of Table~\ref{dataset_parameters}. 20 additional color-corrected composite photos of each plant from the same age range unseen during the training process compose the distribution $V$ for qualitative evaluation of the models. Dataset parameters for this section are listed in Table~\ref{additional_dataset_parameters}.

\begin{table}[H]
\footnotesize
\begin{center}
\begin{tabular}{l c c c c c c c c} \hline
Dataset Name & $N_x$ & $N_y$ & Species & Age (days) & $N_{\mathrm{backgrounds}}$ & $S_{\mathrm{min}}$ & $S_{\mathrm{max}}$ & Figure \\ \hline
Color-Corrected Composites 4 & 64 & 64 & Canola & 10-40 & 32 & 0.50 & 0.85 & \ref{canola} \\ \hline
Color-Corrected Composites 5 & 64 & 64 & Oat & 0-365 & 32 & 0.50 & 0.85 & \ref{oat} \\ \hline
Color-Corrected Composites 6 & 64 & 64 & Wheat & 0-365 & 32 & 0.50 & 0.85 & \ref{wheat} \\ \hline
\end{tabular}
\end{center}
\caption{Parameters for each additional training dataset.}
\label{additional_dataset_parameters}
\end{table}

\begin{figure}[H]
\centering
\includegraphics[width=\linewidth]{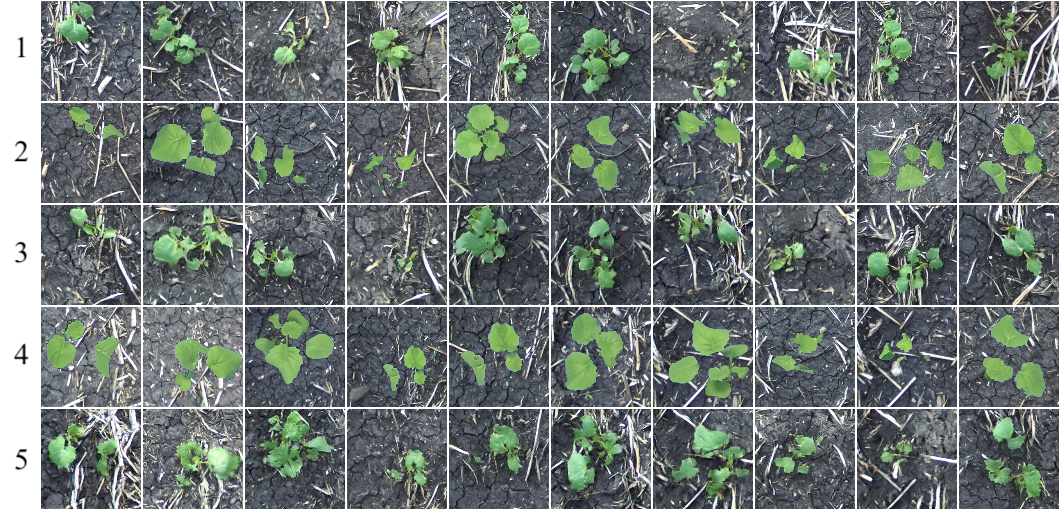}
\caption{Canola images sampled from the distribution $Y$ (1), images sampled from the distribution $X$ (2), translated images $G(\bm{x})$ (3), images sampled from the distribution $V$ (4), and translated images $G(\bm{v})$ (5).}
\label{canola}
\end{figure}

\begin{figure}[H]
\centering
\includegraphics[width=\linewidth]{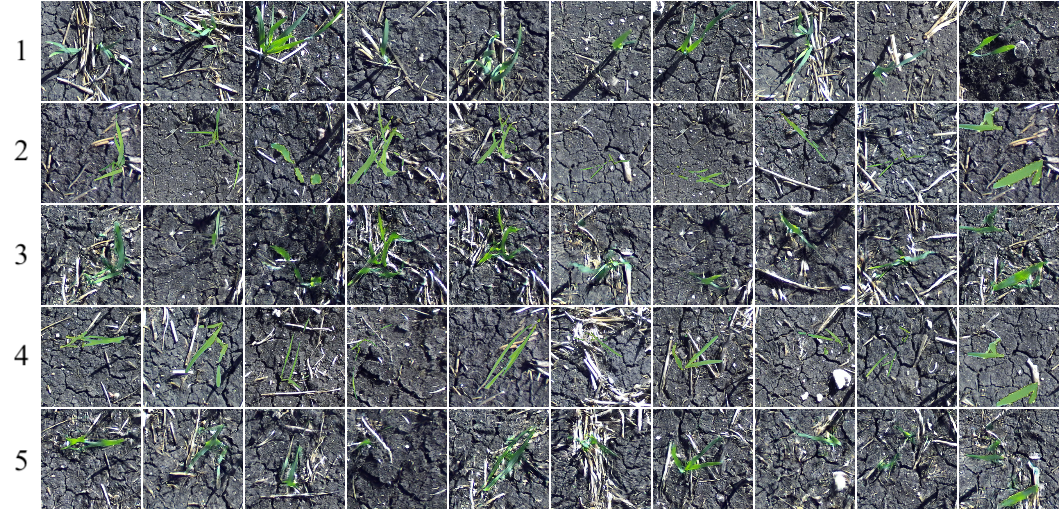}
\caption{Oat images sampled from the distribution $Y$ (1), images sampled from the distribution $X$ (2), translated images $G(\bm{x})$ (3), images sampled from the distribution $V$ (4), and translated images $G(\bm{v})$ (5).}
\label{oat}
\end{figure}

\begin{figure}[H]
\centering
\includegraphics[width=\linewidth]{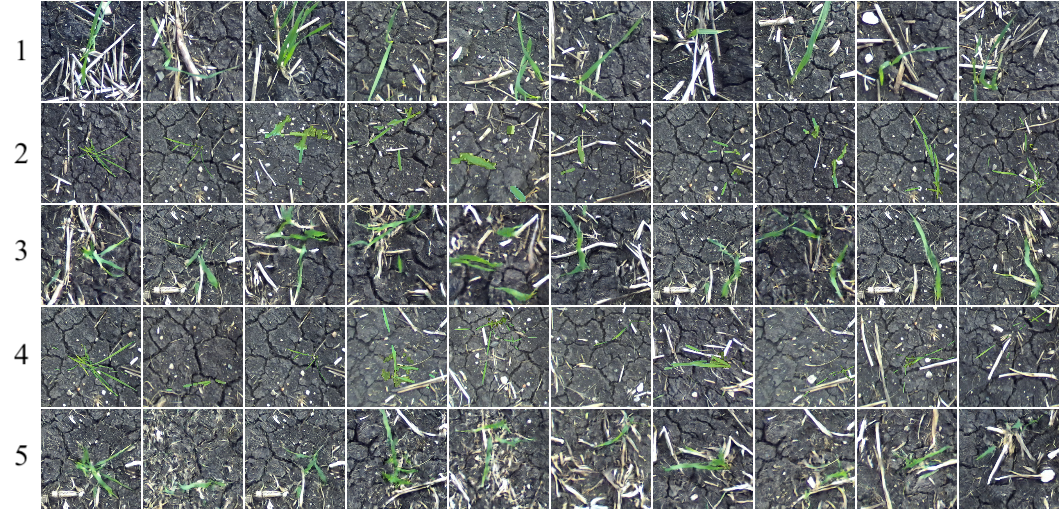}
\caption{Wheat images sampled from the distribution $Y$ (1), images sampled from the distribution $X$ (2), translated images $G(\bm{x})$ (3), images sampled from the distribution $V$ (4), and translated images $G(\bm{v})$ (5).}
\label{wheat}
\end{figure}

\section*{Declaration of Competing Interest}

The authors declare that they have no known competing financial interests or personal relationships that could have appeared to influence the work reported in this paper.

\section*{Acknowledgments}

We gratefully acknowledge the support of the University of Winnipeg, the Natural Sciences and Engineering Research Council of Canada, Compute Canada (now Digital Research Alliance of Canada),  Western Economic Diversification Canada, and Mitacs.


\end{document}